\documentclass{article}

\PassOptionsToPackage{numbers}{natbib}
\usepackage[preprint]{neurips_2024}

\usepackage[preprint]{neurips_2024}

\usepackage{amsmath}
\usepackage[utf8]{inputenc} 
\usepackage[T1]{fontenc}    
\usepackage{hyperref}       
\usepackage{url}            
\usepackage{booktabs}       
\usepackage{amsfonts}       
\usepackage{nicefrac}       
\usepackage{microtype}      
\usepackage{xcolor}         
\usepackage{graphicx}         
\usepackage{caption}
\usepackage{graphicx}
\usepackage{float} 
\usepackage{subcaption}

\title{Bridging the Dimensional Chasm:\\
Uncover Layer-wise Dimensional Reduction in Transformers through Token Correlation}

\author{%
  Zhuo-Yang Song$^{1}$ \quad Zeyu Li$^{2}$ \quad Qing-Hong Cao$^{1,3}$ \quad Ming-xing Luo$^{4}$ \quad Hua Xing Zhu$^{1,3}$ \\
  $^1$School of Physics, Peking University, Beijing 100871, China \\
  $^2$CAS Key Laboratory of Theoretical Physics, Institute of Theoretical Physics,\\ Chinese Academy of Sciences, Beijing 100190, China \\
  $^3$Center for High Energy Physics, Peking University, Beijing 100871, China \\
  $^4$Beijing Computational Science Research Center, Beijing 100193, China \\
}

\begin{document}

\maketitle

\begin{abstract}
The geometric evolution of token representations in large language models (LLMs) presents a fundamental paradox: while human language inherently organizes semantic information in low-dimensional spaces ($\sim 10^1$ dimensions), modern LLMs employ high-dimensional embeddings ($\sim 10^3$ dimensions) processed through Transformer architectures. To resolve this paradox, this work bridges this conceptual gap by developing a geometric framework that tracks token dynamics across Transformers layers. Through layer-wise analysis of intrinsic dimensions across multiple architectures, we reveal an expansion-contraction pattern where tokens diffuse to a "working space" and then progressively project onto lower-dimensional submanifolds. Our finding implies a negative correlation between the working space dimension and parameter-sensitive performance of the LLMs, and indicates that effective models tend to compress tokens into approximately 10-dimensional submanifolds, closely resembling human semantic spaces. This work not only advances LLM interpretability by reframing Transformers layers as projectors that mediate between high-dimensional computation and low-dimensional semantics, but also  provides practical tools for model diagnostics that do not rely on task-specific evaluations.
\end{abstract}

\section{Introduction}

The classic game of "Twenty Questions" vividly illustrates the remarkable efficiency of human language. By posing a series of binary questions, a skilled questioner can swiftly identify any concept through a sequence of yes-or-no responses.
This phenomenon strongly suggests that natural language inherently possesses a low dimensionality, with human linguistic representations likely residing in spaces as compact as $\sim10^1$ dimensions~\cite{idoftext}. 
In stark contrast, modern LLMs embed discrete tokens into high-dimensional vector spaces ($\sim10^3$ dimensions)~\cite{wordvector}, employing complex transformations to generate coherent text~\cite{attentionneed}. 
Despite these models' unprecedented capabilities in language understanding and generation, a fundamental question remains unresolved: How do token representations evolve from high-dimensional embeddings into structures that mirror the low-dimensional organization of natural language?

Current mechanistic interpretations of Transformer primarily view it as interacting particle systems~\cite{mathattention,mathattentionII,mathattentionIII,phase}, where tokens cluster or disperse across different layers. 
However, the geometric properties of these evolving representations, particularly their intrinsic dimensionality and manifold structure, remain underexplored.
While previous research has examined intrinsic dimensions (ID) in neural networks~\cite{dimensionstart,ID,IDhavemeaning,IDofimage,idofmodel,imagedimension}, two critical gaps remain:  
(1) Most analyses focus on prompt-level representations (e.g., final token embeddings~\cite{layers,layersndtoken}), overlooking token-level dynamics within sequences; (2) The relationship between geometric properties (e.g., manifold dimensionality) and model performance remains speculative, with limited empirical grounding.

In this work, we bridge these gaps by introducing a novel geometric framework that analyzes token evolution in LLMs.
The central idea of our work is to introduce the correlators—dynamical quantities that can quantify the interplay between token representations and their underlying manifolds across Transformers layers.
We demonstrate that tokens do not occupy the full $\sim10^3$ dimensional embedding space but instead evolve on significantly lower-dimensional submanifolds. 
By investigating different models, we discover a potentially negative relationship between the intrinsic dimension of model's working manifold and generalization performance~\cite{Qwenbench,QwenbenchII,deepseekdistill}.

Our contributions are as follows:

(1) Token-Level Geometric Analysis:
We track the intrinsic dimension (ID) of token representations layer-by-layer, uncovering a non-monotonic "dimensional expansion-contraction" pattern that is independent of the input sentence structure.

(2) Correlators as Dynamical Probes:
We introduce the concept of "correlators" as an interpretable metric that bridges the geometric properties of the model's internal representations with its overall behavior. This metric elucidates how token representations within the model transition from high-dimensional spaces to lower-dimensional subspaces across successive layers. This transition reveals the progression of models from a high-dimensional machine "working space," rich with complex computational features, towards a lower-dimensional "semantic space" that closely resembles human semantic representation~\cite{wordtrack,layersndtoken,peakII}. The correlator we introduce is akin to cosine similarity, but it also incorporates the radial information of a vector. In certain limits, the correlator tends to align with cosine similarity.

(3) Performance-Dimensionality Trade-off:
Upon analyzing various LLMs, particularly the Qwen2.5 series~\cite{Qwenbench,QwenbenchII}, we have identified a potential relationship between the dimensionality of the model's operational space, denoted as $d_{\text{model}}$, and its performance across different tasks. Notably, we find that models operating in lower-dimensional spaces tend to exhibit superior performance. This correlation suggests that reducing $d_{\text{{model}}}$ could enhance task efficiency and effectiveness. Our findings provide a practical and rapid criterion for assessing the potential of LLMs. By examining the dimensionality of the model's operational space, we can gain insights into its likely performance without the need for extensive task-specific evaluations.

The remainder of this paper is organized as follows: Section~\ref{gen_inst} introduces the related work. Section~\ref{section_method} presents the geometric projection theory and the method for obtaining the model dimensions. Section~\ref{sec:exp} describes the experimental results on Qwen~\cite{Qwenbench,QwenbenchII,deepseekdistill}, Mistral~\cite{mistral7b}, and Llama~\cite{llama3herdmodels}. Finally, Section~\ref{sec:conclusion} summarizes our work.

This study aims to enhance our understanding of the internal workings of LLMs by reinterpreting Transformers layers as dimensional projectors. It also offers potential tools for assessing model capabilities based on geometric signatures.

\section{Related Work}
\label{gen_inst}

The concept of intrinsic dimensionality has been widely studied to enhance our understanding of neural representations~\cite{attentiondimension,dimensionstart,idofmodel,layers,layersndtoken,imagedimension,phase,topologicalframe,wordtrack}. Early studies~\cite{dimensionstart} demonstrated that high-dimensional problems often exhibit low intrinsic dimensionality, as explained by random projection theory. This phenomenon has been further confirmed in various domains~\cite{imagedimension,IDofimage} and language model analyses ~\cite{IDhavemeaning,phase}. 

Notably, recent investigations into LLMs suggest that human language may inherently reside in a low-dimensional semantic manifold, despite the combinatorial complexity and vast vocabulary of natural language. These studies demonstrate that semantic relationships can be efficiently encoded in surprisingly low-dimensional spaces~\cite{canbedistinguish,idoftext}, akin to the manner in which image manifolds compress visual information.

Building on these findings, recent advancements have proposed distinct operational definitions of intrinsic dimensionality with different scales:$\sim 10^2$~\cite{idofmodel}, which correlates with the relationships between parameters shaped by pre-training, and $\sim 10^1$~\cite{imagedimension,phase}, a semantic-space dimensionality tied to model architecture.
While existing estimators (e.g.,~\cite{attentiondimension,topologicalframe}) provide valuable methodological foundations, our work distinguishes itself by explicitly decoupling these two dimensions and linking them to model dynamics.

Prior studies on layer-wise representations (e.g.,~\cite{layers}) indicates that inter-layer propagation can modify token structures according to specific laws ~\cite{mathattentionII,mathattentionIII}, which aligns with our findings.
Our contribution extends these findings by introducing correlators as quantitative descriptors  of dimension-specific signal propagation, moving beyond static dimensionality measurements or simplified models ~\cite{layersndtoken,mathattention,mathattentionII,mathattentionIII}.

While attention mechanisms are known to influence model performance ~\cite{attentionneed,mathattention,mathattentionII,mathattentionIII}, our analysis reveals more complex relationships within specialized LLMs. Specifically, we found that a lower value of $d_{\text{model}}$ (even when the architecture remains fixed) correlates with improved task performance. In contrast, $d_{\text{machine}}$ seems to be limited by the design of the architecture. This finding partially aligns with superficial similarity metrics ~\cite{phase} but underscores the need for deeper investigation. Our framework tentatively explains this phenomenon through the concept of correlator stability across different layers, presenting a mechanism that is distinct from previous topological analyses ~\cite{topologicalframe,idoftext}.

\section{Method}\label{section_method}
In this section, we introduce the core methodology of this paper, focusing on the "projection process" within the Transformer architecture.
We demonstrate that an approximately conserved quantity exists during the forward propagation of Transformers. This quantity serves as a foundation for calculating the intrinsic dimension of the model and the dimension of the semantic space learned by the model.

\subsection{Transformer as an operator}
All modern LLMs are based on Transformer architecture, which consists of multiple layers stacked on top of each other. Each layer includes components such as self-attention mechanisms, positional encoding and feed-forward network layers etc. We will regrad each of the layer as a fundamental unit to study the working mechanism of Transformers.

\begin{figure}[t]
\centerline{\includegraphics[width=0.8\linewidth]{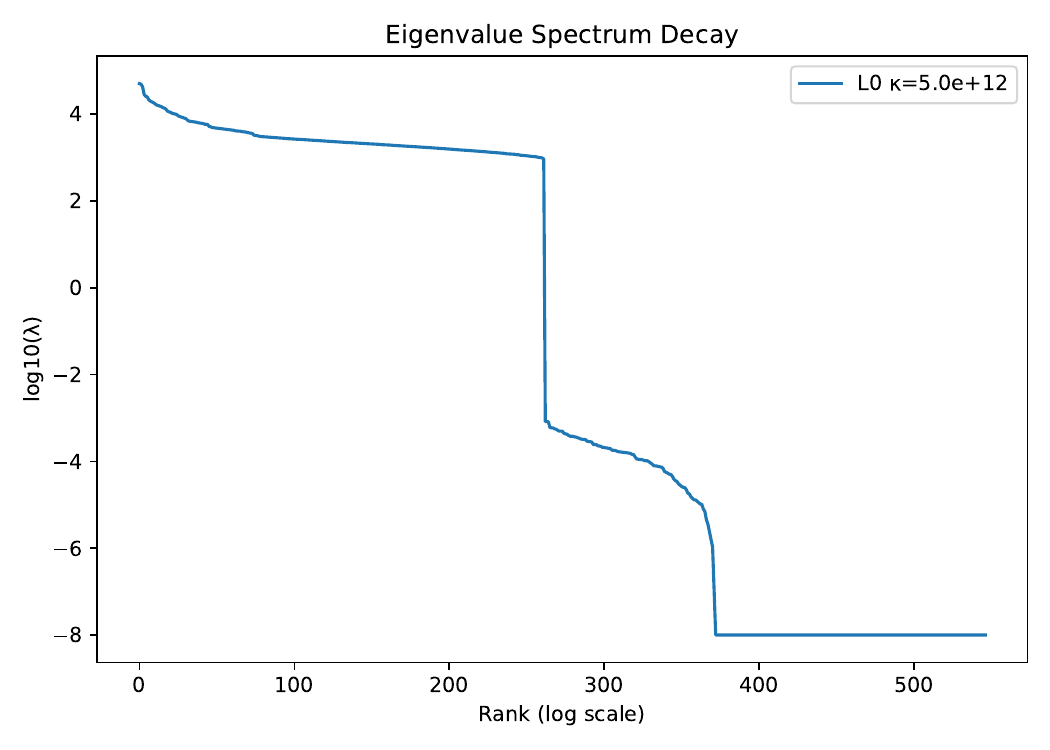}} 
	\caption{Spectrum for sentence sample 1 in Appendix~\ref{subsec:1}. We clip eigenvalues whose values are less than $10^{-8}$ and set them to $10^{-8}$.} \label{fig:eigen_spectrum_Qwen2.5-3B}
\end{figure}

Assume there is a sentence denoted as $\{ x_i\}$. Each $x_i$ corresponds to a word id in this sentence. When this sentence is input into a LLM with $m$ layers, each $x_i$ will first be embedded into a word vector in a high-dimensional space:
\begin{equation}
    {\rm Emb} : \quad x_i \to \boldsymbol{t}_i \in \mathbb{R}^{d_{\text{embed}}}
    \label{eq:x->t}\;,
\end{equation}
where $d_{\text{embed}}$ is set up from the beginning. However, as the word of nature language contains some intrinsic structure, word vector should actually live on a submanifold:
\begin{equation}
    \boldsymbol t_i \in \mathcal{E}_0 \subset \mathbb{R}^{d_{\text{embed}}}\;.
\end{equation}
The high-dimensional word embeddings in LLMs (typically ($d \sim 10^3$) result in ill-conditioned Gram matrices ($G_{ij} = \boldsymbol t_i\cdot \boldsymbol t_j$) with extreme condition numbers ($\kappa(\mathcal{E}) \equiv \lambda_{max}(G)/\lambda_{min}(G)\gg 1$).
This indicates nearly linear-dependent basis vectors, which contradicts the desired compactness of semantic space. 
For instance, we compute the Gram matrices for sentence sample 1 in Appendix~\ref{subsec:1} by Qwen2.5-3B and present its spectrum in Fig.~\ref{fig:eigen_spectrum_Qwen2.5-3B}. We can see the eigenvalue spectrum is not continuously distributed and display a pathological distribution~\cite{rank,ALBERT}.

Through successive Transformers layers, the ill-conditioned embedding space
$\mathcal{E}_0$ will progressively be refined by attention mechanisms~\cite{attentionneed}, which reduce linear dependencies by suppressing spurious correlations. Such optimization compresses $\mathcal{E}_0$ into a low-dimensional intrinsic manifold $\mathcal{E}_\text{machine}$, which aligns with the compact semantic structure of human language~\cite{idoftext}. This process resolves the issues of pathological geometry while maintaining linguistic coherence.

Within this framework, we formalize the effect of each Transformer layer as an operator:
\begin{equation}
    \mathcal{T}_{\xi}: \quad \mathcal{E}_{\xi} \to \mathcal{E}_{\xi+1}\;,
\end{equation}
where \(\xi\) represents the layer number, and \(\mathcal{E}_{\xi}\) denotes the manifold of word vectors after \(\xi\) layers. 
Importantly, the sequence \(\{\mathcal{T}_{\xi}\}\) governs the evolution \(\mathcal{E}_0 \rightarrow \mathcal{E}_m\), with some empirical evidence suggesting universal geometric regularities (e.g., monotonic spectral decay, entropy contraction) across architectures, even though the specific trajectories may vary based on parameters.

To intuitively grasp the nature of \(\mathcal{T}_\xi\), we examine the architecture of Qwen2ForCausalLM~\cite{QwenbenchII,attentionneed}, which implements the operator as follows:
\begin{equation}
    \boldsymbol t_{i,\xi} \; \rightarrow \; \mathcal{T}_\xi \boldsymbol t_{i,\xi} = \boldsymbol t_{i,\xi} + F_\xi\left( \sum_{j,k,l}\hat{\Gamma}_{ijkl,\xi}\left(\hat{\boldsymbol t}_{j,\xi} \otimes \hat{\boldsymbol t}_{k,\xi}\right) \cdot \hat{\boldsymbol t}_{l,\xi} , \; \phi(i) \right)\;,
    \label{eq:transformer_flow}
\end{equation}
where 
\begin{itemize}
    \item \(\hat{\Gamma}_{ijkl,\xi}\) is a fourth-order tensor ,each element represent for a non-linear function, parameterized the  the query-key-value projections at layer \(\xi\). Hat represents for normalization.
    \item \(\hat{\boldsymbol t}_{i,\xi}\) denote normalized token representations at position \(i\) in layer $\xi$.
    \item \({\boldsymbol t}_{i,\xi}\) denote token representations at position \(i\) in layer $\xi$.
    \item \(\otimes\) represents the outer product, which models the correlations between pairs of tokens.
    \item \(\phi(i)\) introduces positional information through position embeddings.
    \item \(F_\xi\) is a nonlinear projection (e.g., MLP) with layer-dependent parameters.
\end{itemize}

This formulation can be generalized to standard Transformers, where \(\mathcal{T}_\xi\) iteratively suppresses spurious geometric modes in \(\mathcal{E}_\xi\) via attention-based reweighting and nonlinear filtering. 
This process drives condition number \(\kappa(\mathcal{E}_\xi) \sim 1\) as \(\xi\) increases. The parameterized operator $\mathcal{T}_\xi$ plays a central role in disentangling semantic dependencies while preserving linguistic invariance~\cite{attentionneed,differentarc}.

In practice, this formula is straightforwardly used to compute the evolution of word vectors. However, it is difficult to describe the evolution of manifold $\mathcal{E}_{\xi}$. Therefore, to intuitively understand $\mathcal{T}_\xi$, we approximate the above procedure as a projection onto the word vector if $\dim (\mathcal{E}_\xi)>\dim(\mathcal{E}_{\xi+1})$:
\begin{align}
    \boldsymbol t_{i,\xi} \; \rightarrow \; \boldsymbol{t}_{i,\xi+1} =  \mathcal{P}_\xi\boldsymbol t_{i,\xi} =  \boldsymbol t_{i,\xi} -\|\boldsymbol t_{i,\xi}\|\sum_\alpha\left((\hat{\boldsymbol t}_{i,\xi}\cdot\boldsymbol{n}_\alpha(\hat{\boldsymbol t}_{i,\xi}))\boldsymbol{n}_\alpha(\hat{\boldsymbol t}_{i,\xi})\right)\;,
    \label{eq:P}
\end{align}
where ${\boldsymbol{n}_\alpha}(\boldsymbol{\hat{t}})$ are normal vectors for manifold $\mathcal{E}_{\xi}$ at the end point of $\boldsymbol{\hat{t}}$, they depend on the input sequence $\mathcal{B}$ and $\alpha=1,\cdots,\dim(\mathcal{E}_{\xi})-\dim(\mathcal{E}_{\xi+1})$. 
The projection operation $\mathcal{P}_\xi$ means tokens will ultimately cluster together, and this agree with previous work~\cite{mathattentionII,cluster}. Such operation will eliminate certain freedoms of the word vectors and converge the word vectors within the same batch towards a target manifold. Because of the complexity of the practical $\mathcal{T}_\xi$, we assume that ${\boldsymbol n_\alpha}(\boldsymbol{\hat{t}})$ satisfies :
\begin{equation}\label{n_sta}
    \langle \boldsymbol{n}_\alpha(\boldsymbol{\hat{t}}_i) \cdot \boldsymbol{t}_j\rangle=\langle \boldsymbol{n}_\alpha(\boldsymbol{\hat{t}}_i) \cdot \boldsymbol{n}_\beta(\boldsymbol{\hat{t}}_j)\rangle=0,\quad  \forall \; i , j\,,
\end{equation}
where the $\langle \cdots\rangle$ represents taking the average over the input sequence $\mathcal{B}$. For example, $\langle \|t_k\|^2 \rangle = \frac{\sum_{\boldsymbol t_k  \in \mathcal{B}}{\|\boldsymbol t_k\|^2}}{\sum_{\boldsymbol t_k \in \mathcal{B}}{1}}$, $ \langle \boldsymbol t_i\cdot \boldsymbol t_j \rangle = \frac{\sum_{\boldsymbol t_i,\boldsymbol t_j \in \mathcal{B}}{\boldsymbol t_i\cdot \boldsymbol t_j}}{\sum_{\boldsymbol t_i,\boldsymbol t_j \in \mathcal{B}}{1}}$.
 
However, the high complexity of the manifold and evolution process makes it impractical to explicitly write down the exact form of $\boldsymbol n_\alpha(\boldsymbol{\hat{t}})$ and perform specific calculations using eq.~\eqref{eq:P}. Therefore, we need to find a new quantity that both characterizes the evolution of the manifold and is easy to calculate.

\subsection{The correlator of word vector}
We first examine the process of semantic formation. Suppose the word vector passing through $\xi$ layers is $\boldsymbol t_{i,\xi}$, we input this vector into the $\text{lm\_head}$ layer and investigate its output. Take Qwen2.5-3B model as an example, we input sentence sample 1 in Appendix~\ref{appendix:wiki_samples} and collect all outputs of $\text{lm\_head}$ layer for $t_{i,\xi}$, as shown in Tab.~\ref{tab:token}.
We can see that as the layers become deepen, the outputs of $\text{lm\_head}$ layer first become chaotic, then gradually become ordered, ultimately conforming to words that are consistent with semantics and logic~\cite{wordtrack}.

This phenomenon inspire us that a natural way to describe the distribution of these word vectors is their $\text{token correlator}$ (for simplicity, we will call it correlator below):
\begin{equation}
E(\xi) = \frac{\sum_{i,j}{(\boldsymbol t_{i,\xi} \cdot \boldsymbol t_{j,\xi}) }}{Len(\mathcal{B})\sum_k{
\|\boldsymbol t_{k,\xi}\|^2 }} = \frac{\langle \boldsymbol t_{i,\xi} \cdot \boldsymbol t_{i,\xi}\rangle}{\langle
\|\boldsymbol t_{i,\xi}\|^2\rangle }; \quad \boldsymbol t_i,\boldsymbol t_j,\boldsymbol t_k\in \mathcal{B}, \quad i \neq j
\label{eq:correlator_def}.
\end{equation}

For each layer $\xi$, we compute the correlator $E(\xi)$ by averaging the pairwise dot products of token vectors within a batch $\mathcal{B}$ and normalize them by their squared norms.  

The correlator will tend to 0 if the vectors are randomly distributed and may increase when transformed to a lower dimension manifold. This aligns with the particle view as derived by previous work~\cite{mathattentionIII}.

\begin{figure}[t]
    \centering
    \begin{subfigure}[t]{1\textwidth}
        \centering
        \includegraphics[height=4.5cm]{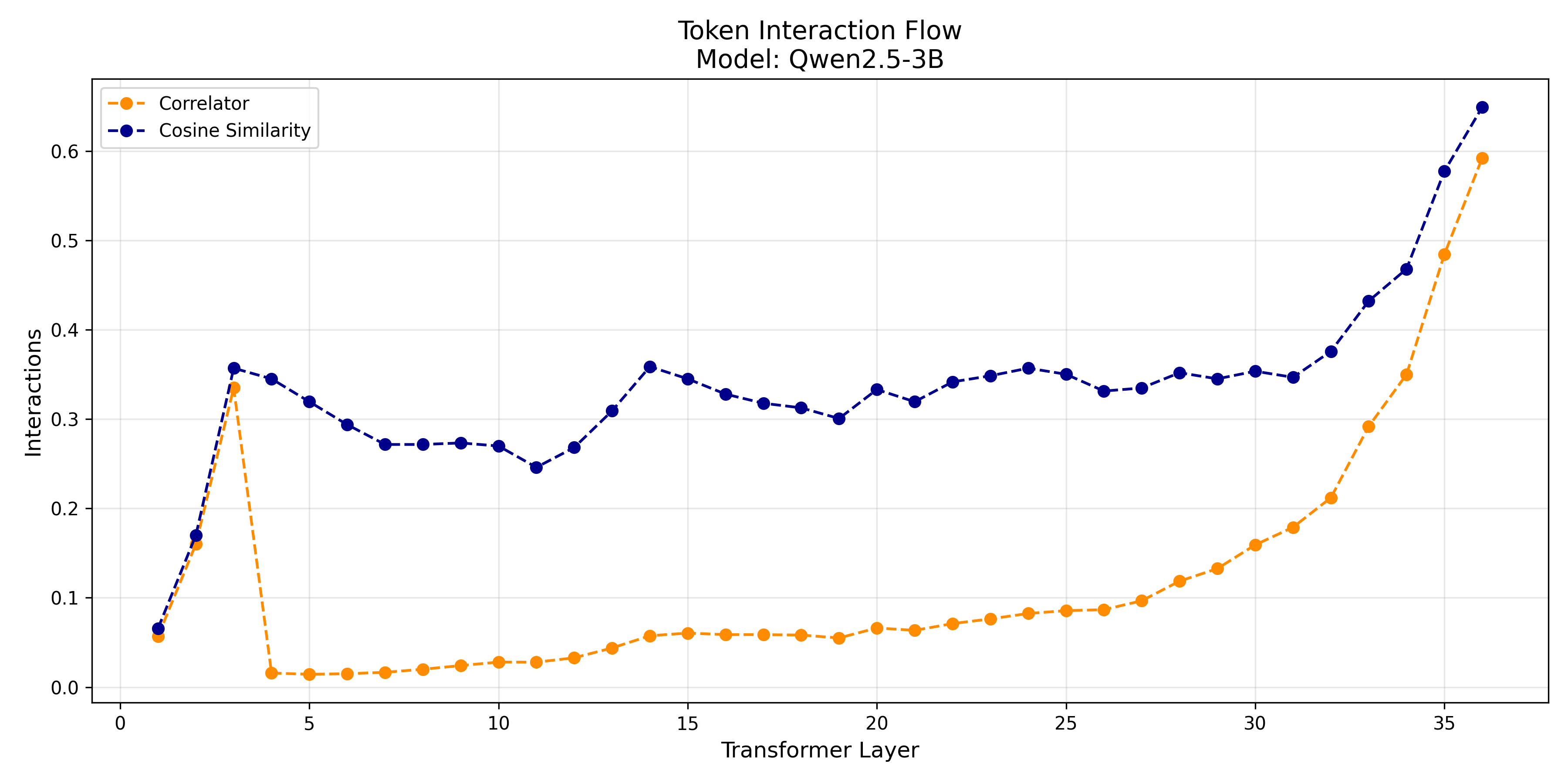}
        \caption{Input length 547 tokens}
        \label{fig:sub1}
    \end{subfigure}
    \vfill 
    \begin{subfigure}[t]{1\textwidth}
        \centering
        \includegraphics[height=4.5cm]{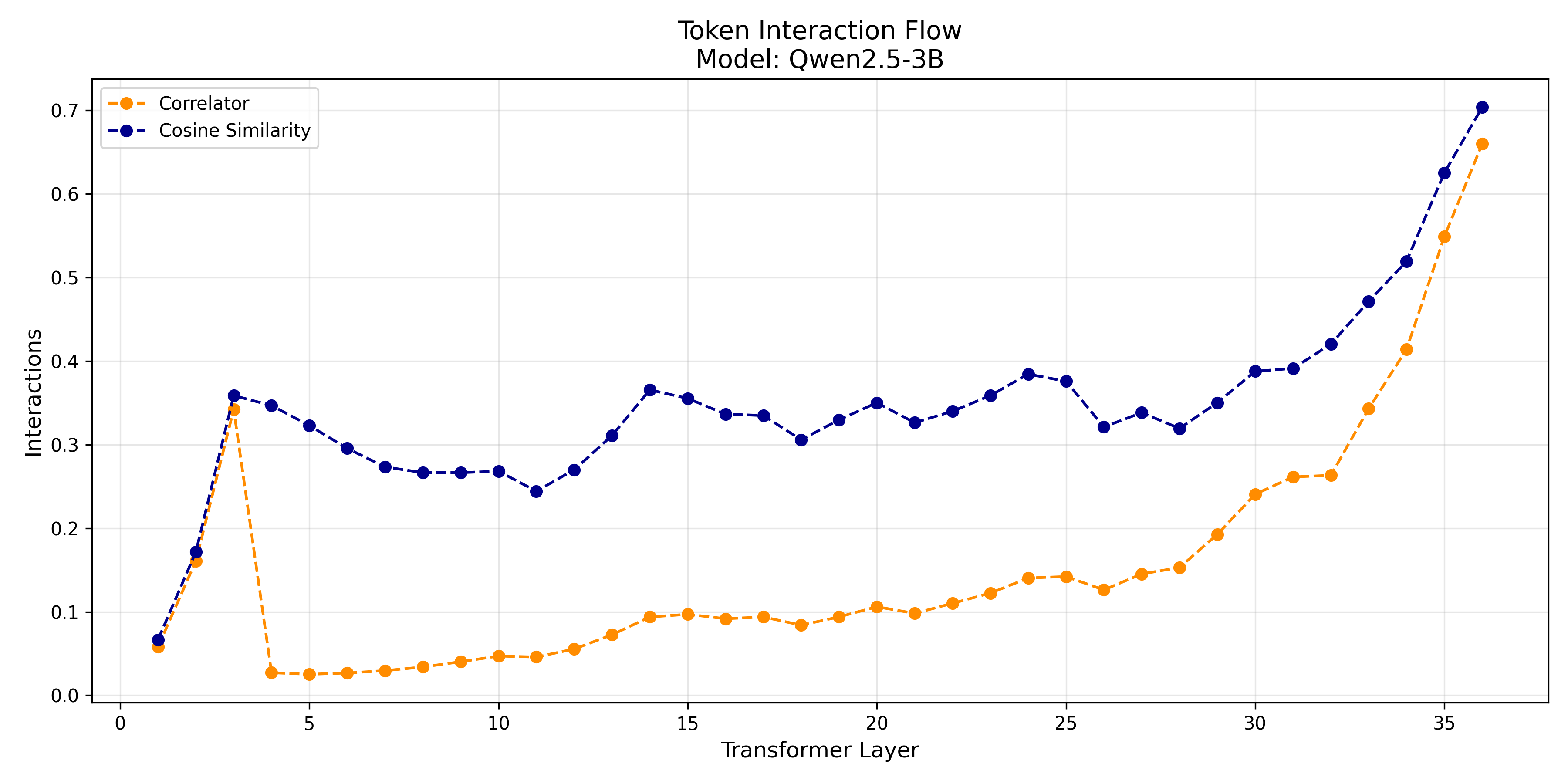}
        \caption{Input length 1094 tokens}
        \label{fig:sub2}
    \end{subfigure}
    \vfill 
    \begin{subfigure}[t]{1\textwidth}
        \centering
        \includegraphics[height=4.5cm]{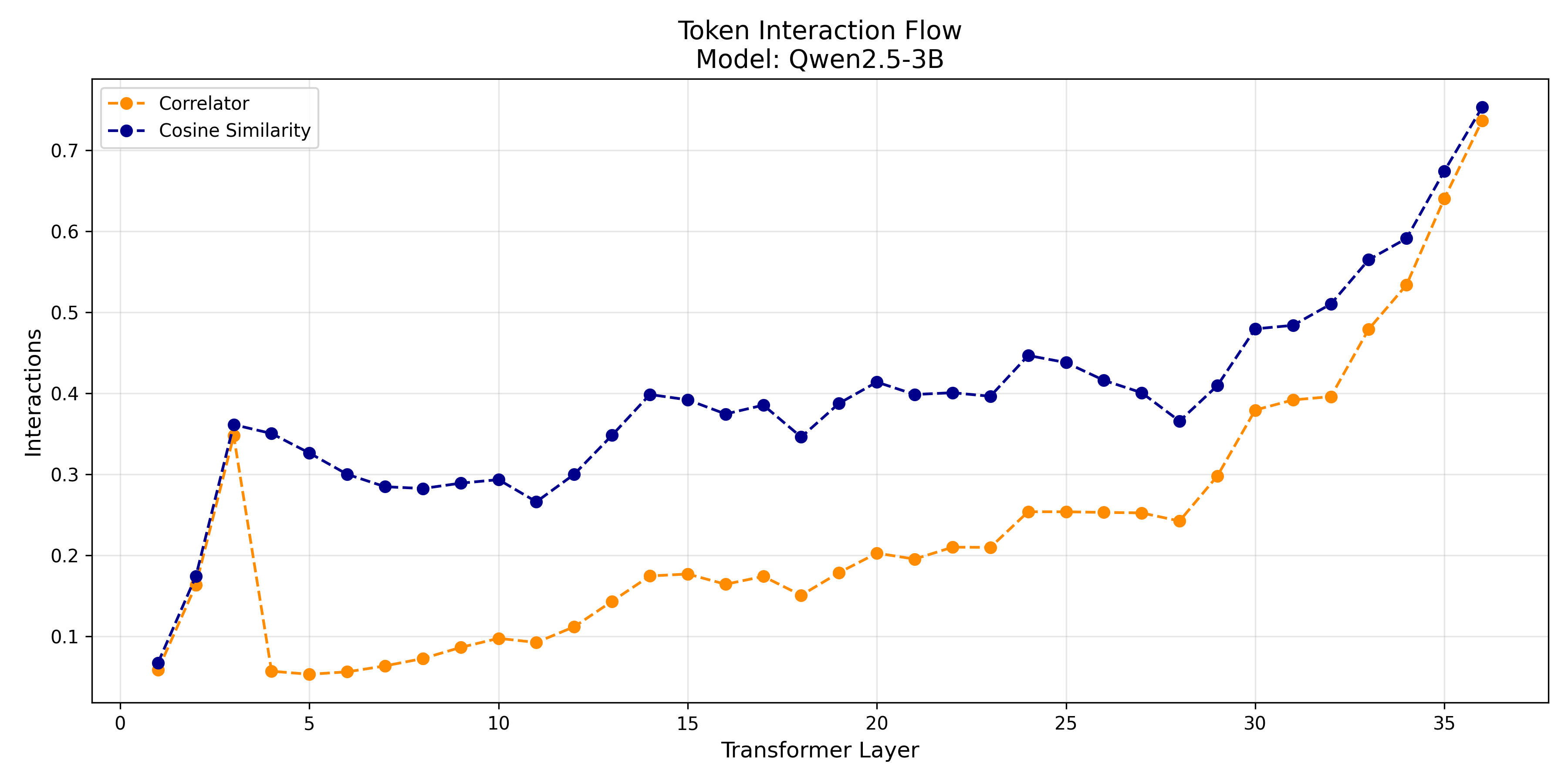}
        \caption{Input length 2735 tokens}
        \label{fig:sub3}
    \end{subfigure}
    \caption{How the correlator changes in each layer. The trends of the correlator and Cosine Similarity are consistent.}
    \label{fig:Qwen2.5-3B}
\end{figure}

As a concrete example, we compute the correlator of the word vectors in each layer and draw them in Fig.~\ref{fig:Qwen2.5-3B}. The data reveal that the correlator initially reaches a peak, then quickly decline into a small value which is very close to 0. Subsequently, the correlator gradually increase and approach to $\sim1$, which agrees with analysis in previous work~\cite{mathattentionII,cluster}.

In fact, when we seek for a normalized correlation, only two definitions are sufficiently symmetrical: the correlator itself (eq.~\eqref{eq:correlator_def}) and cosine similarity. Although these two methods are theoretically identical, they do have practical differences, mainly due to variations in the length of the input text. However, both methods demonstrate similar trends and essentially show the same input-output relationships as illustrated in Fig.~\ref{fig:Qwen2.5-3B}.

Although both normalized correlators and cosine similarities provide symmetric measures of vector alignment, we intentionally select the correlator for the below analysis. 
This choice stems from the correlator's preservation of radial distribution characteristics: when token vectors evolve across layers, the correlator could capture the changes in radial distribution. This property enables our framework to holistically capture the distributional patterns in the full vector space rather than being constrained to angular relationships on the unit hypersphere.

\begin{table}
    \centering
    \begin{tabular}{cc}
        \hline
        Layer0 & " ent" "ang" "lement" "." "You" \\
        \hline
        Layer5 & ".bootstrap" "queryString" " scrollTo" " fkk" " yourself" \\
        \hline
        Layer10 & "<?> density" "lore" ".compat" "<?> since the 18th National Congress" " yourself" \\
        \hline
        Layer15 & "BIG" "<?> connection" "swer" "<?>" " Majesty" \\
        \hline
        Layer20 & "acular" "<?> UIG" "<?>" " yourself" \\
        \hline
        Layer25 & "<?> concept" "<?> in our country" " :)" " yourself" \\
        \hline
        Layer30 & "<?> enlightenment" "<?>" "Answer" " yourself" \\
        \hline
        Layer36 & "ang" "lement" " and" "You" " are" \\
        \hline
    \end{tabular}
    \caption{The table illustrates the layer-by-layer output of Qwen2.5-3B~\cite{wordtrack} where ``<?>'' represents chinese words. It demonstrates the transition from the ID space in the initial layers to a machine language space (decode at T=0), where word vectors can be decoded but lack human semantic meaning. As the layers progress towards the final output, meaningful words begin to emerge, culminating in human-readable language at the final layer. This table captures the transformation process within the Transformer architecture as it evolves from ID space to machine language and ultimately to human language.}
    \label{tab:token}
\end{table}

Therefore, we conjecture that the projection process of Transformers can be characterized by two distinct stages. In the initial 10\% of the layers, word vectors gradually diffuse into a higher-dimensional manifold, denoted as $\mathcal{E}^*$. We define this manifold, $\mathcal{E}^*$, as the machine language manifold and the model's working space. This transition aligns with previous findings~\cite{Resnet,peakII}, where the model expands the initial embedding space to capture richer features. Subsequently, the model begins to project the word vectors into a lower-dimensional manifold, as described in eq.~\eqref{eq:P}. This second stage transitions the model from its working space to a semantic space that closely resembles human language. The exponential behavior observed in this process can be partly explained by theories from prior research~\cite{mathattentionIII}.

We argue that the core issue with the working space \(\mathcal{E}^*\) lies in the fact that the attention mechanism is not fully utilized. This limitation results in distortion for a lower-dimensional space and causes the two values (correlators and cosine similarity) to diverge. In other words, the model must transition from the input space to an equivalent higher-dimensional space to acquire sufficient features for further evolution. The extent to which dimensionality is increased is negatively related to the model’s capability or the completeness of the text. This relationship highlights the significance of the subsequent study of the correlator $E$.

The final output asymptotically approaches 1 but never reaches it. This is because word vectors reside on the semantic manifold $\mathcal{E}_{\text{machine}}$ learned by the Transformers, which we hypothesize aligns with the human's intrinsic semantic space $\mathcal{E}_{\text{int}}$ 
described in previous work~\cite{idoftext}. 
This structural constraint ensures that successive token predictions remain proximate within the semantic space. 
However, achieving perfect uniformity ($E=1$) would imply identical predictions across all tokens. This scenario represent a degenerate solution that cannot correspond to an optimal configuration or even a meaningful local minimum in the optimization landscape~\cite{oversmooth,oversmoothII}.

\subsection{Dynamic of correlator}
Following the concept of manifold evolution and projection of word vectors, we establish the dynamics of the correlator.

We investigate the changes of correlator after word vectors passing through a certain layer. As mentioned in last subsections, a word vector passing through $\xi$ layers can be represented as eq.~\eqref{eq:P}.
Then the correlator $E(\xi+1)$ is:
\begin{align}\label{eq:E_xi+1}
    E(\xi+1) &= \frac{\langle \boldsymbol{t}_{i,\xi+1} \cdot \boldsymbol t_{j,\xi+1}\rangle}{\langle{\|\boldsymbol t_{k,\xi+1}\|^2 }\rangle} \nonumber \\ &= \frac{\langle{\boldsymbol t_{i,\xi} \cdot \boldsymbol t_{j,\xi}}\rangle}{\langle  \| \boldsymbol t_{k,\xi} -\|\boldsymbol t_{k,\xi}\|\sum_\alpha\left((\boldsymbol{\hat{t}}_{k,\xi} \cdot\boldsymbol{n}_\alpha) \boldsymbol{n}_\alpha\right)  \|^2 \rangle} \nonumber \\
    &= \frac{\langle \boldsymbol t_{i,\xi} \cdot \boldsymbol t_{j,\xi}\rangle}{\langle\|\boldsymbol t_{k,\xi}\|^2\rangle} \cdot\frac{1}{1 -\langle \sum_{\alpha}(\boldsymbol{\hat{t}}_{k, \xi} \cdot \boldsymbol{n}_{\alpha})^2 \rangle} \nonumber \\
    &= E(\xi) \cdot \frac{1}{1 -\langle \sum_{\alpha}(\boldsymbol{\hat{t}}_{k, \xi} \cdot \boldsymbol{n}_{\alpha})^2 \rangle}\,.
\end{align}
In the second equality, we use the statistical properties for $\vec{n}_{\alpha}$ in eq.~\eqref{n_sta}, and in the third equality we assume:
\begin{equation}
    \langle ||\boldsymbol t_{k,\xi}||^2(1-\sum_{\alpha}(\boldsymbol {\hat{t}}_{k, \xi} \cdot \boldsymbol{n}_{\alpha}(\boldsymbol {\hat{t}}_{k,\xi}))^2)\rangle = \langle||\boldsymbol  t_{k,\xi}||^2 \rangle \langle 1-\sum_{\alpha}(\boldsymbol {\hat{t}}_{k, \xi} \cdot \boldsymbol {n}_{\alpha}(\boldsymbol {\hat{t}}_{k,\xi}))^2\rangle\,.
\end{equation}

Then all we need is to calculate the $\langle \sum_{\alpha}(\boldsymbol{\hat{t}}_{k, \xi} \cdot \boldsymbol{n}_{\alpha})^2 \rangle = \langle \sum_{\alpha}\cos^{2}{\theta_{k,\alpha}}   \rangle = \sum_{\alpha}\langle \cos^{2}{\theta_{k,\alpha}} \rangle $. We leave the detailed derivation process in Appendix~\ref{appendix:d}, and provide the result:
\begin{equation}
    \sum_{\alpha}\langle \cos^{2}{\theta_{k,\alpha}} \rangle = -\frac{\Delta d}{d-1}
    \label{eq:averaged}\,,
\end{equation}
where $d=\dim(\mathcal{E}_{\xi})$ and $\Delta d = \dim(\mathcal{E}_{\xi+1}) - \dim(\mathcal{E}_{\xi})$. So we can further express the eq.~\eqref{eq:E_xi+1} as:
\begin{equation}
    E(\xi+1) = E(\xi)\left(\frac{d-1}{d-1+\Delta d}\right)\,.
\end{equation}
Thus the difference between the $E(\xi+1)$ and $E(\xi)$ is:
\begin{equation}
    E(\xi+1) - E(\xi) = \Delta E =  -E(\xi) \left(\frac{\Delta d}{ d-1 +\Delta d}\right) \approx -E(\xi) \frac{\Delta d}{d - 1}\,.
\end{equation}
By solving this difference equation, we obtain:
\begin{equation}\label{eq: Ed=c}
    E(\xi) \cdot (d-1) = \text{Constant}\,,
\end{equation}
and this tell us that the product of correlator and the dimension should approximately be a constant. Recall that the projector is no more than an approximation, we may choose eq.~\eqref{eq: Ed=c} as a basic principle.

\begin{figure}
    \centering
    \includegraphics[width=0.6\linewidth]{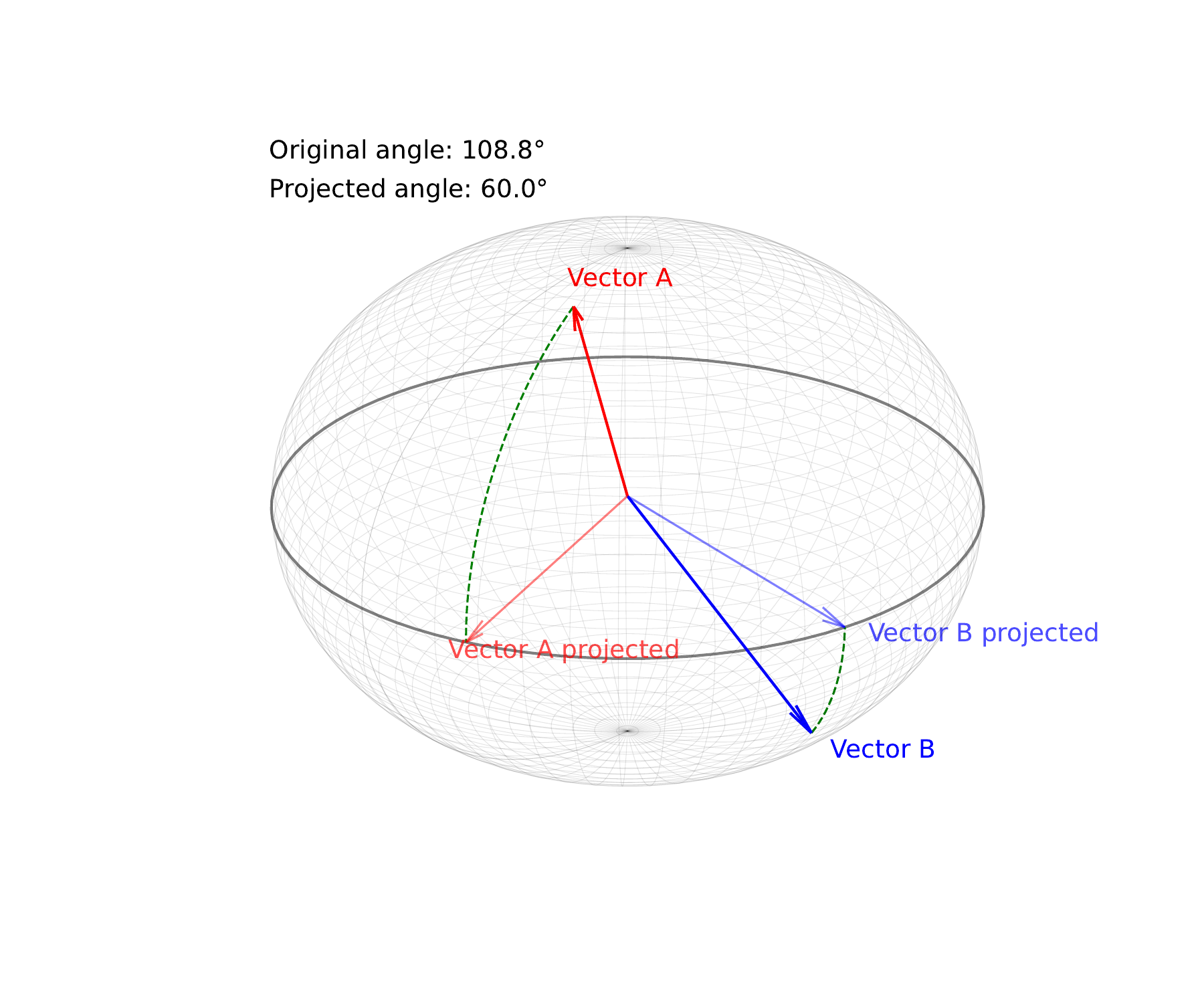}
    \caption{Schematic Diagram of Projection. This figure illustrates how the projection of word vectors affects the angles between them.}
    \label{fig:projector}
\end{figure}

An intuitive conclusion can be drawn from Fig.~\ref{fig:projector}: when a dimension is compressed during the projection process, the correlator between the vectors tends to increase gradually with the projection. 
This phenomenon indicates that the relationships between vectors, which may not be very pronounced in the original high-dimensional space, can become more apparent in the lower-dimensional projected space.

With the relationship between the correlator and the intrinsic dimension of the manifold established, we can revisit the two stages of models that we have previously discussed in ~\ref{2_stages}. At the beginning of the forward propagation, the correlator decreases, which means that the word vector given by eq.~\eqref{eq:x->t} diffuse into a higher dimension manifold, $\mathcal{E}^{*}$. 
This manifold, $\mathcal{E}^{*}$, constitutes the actual working space of the Transformers
and we conjecture its dimension $d_{\text{model}}$ represents the intrinsic dimension of the Transformers. 
On the other hand, the correlator grows rapidly at the second stage, indicating that the manifold is converging to the low-dimension semantic manifold with dimension $d_{\text{machine}}$.

Using eq.~\eqref{eq: Ed=c}, we have:
\begin{equation}
    E_{\text{model}} (d_{\text{model}}-1) = E_{\text{machine}} (d_{\text{machine}}-1) = \text{Constant}\,,
\end{equation}
where $E_{\text{model}}$ is the minimum value of correlator within a forward propagation and $E_{\text{machine}}$ is the correlator of the final layer.
So as long as we can determine the common constant, we can extract $d_{\text{model}}$ and $d_{\text{machine}}$ together.

\begin{table}
    \centering
    \begin{tabular}{|c|c|c|l|}
        \hline
         Name&  Space Symbol& Dimensional Symbol&Correlator Symbol \\
         \hline
         Initial space&  $\mathcal{E}_0$&  not mentioned&$E(0)\quad(\xi=0)$\\
         \hline
         machine's semantic space&  $\mathcal{E}_m$&  $d_{\text{machine}}$&$E_{\text{machine}}=E(m)$\\
         \hline
         machine's working space&  $\mathcal{E}^*$&  $d_{\text{model}}$&$E_{\text{model}}=min{(E(\xi))}$\\
         \hline
 Embedding space& $\mathbb{R}^{d_{\text{embed}}}$& $d_{\text{embed}}$&$E_{\text{random}}=E(m)\quad(\text{random weight})$\\
 \hline
    \end{tabular} 
    \vspace{0.2cm}
    \caption{Comparison of some special spaces and their corresponding correlator and dimensional symbols}
    \label{tab:compare}
\end{table}

Among the set of correlators and dimensions, certain ones hold unique significance. We have curated a selection of these, which are detailed in Table \ref{tab:compare} above. As previously discussed, the product of the correlator and the manifold's dimension remains constant. We will leverage this property to calculate the aforementioned dimension.

In order to calculate these dimensions concretely, we first need to determine the constant at a specific point. 
To achieve this, we utilize the predefined parameter $d_{\text{embed}}$. 
As long as we can uniformly and randomly diffuse the word vectors into the $\mathbb{R}^{d_{\text{embed}}}$ space through the aforementioned operations, we can utilize the known $d_{\text{embed}}$ to calculate this constant for each Transformer. 

To achieve this, we randomly initialized the parameters of a pre-trained Transformer, resulting in a completely random, untrained model. 
The word vectors input into such model will randomly diffuse into the expected $\mathbb{R}^{d_{\text{embed}}}$ space. Once again, we use the sentence sample in Appendix~\ref{appendix:wiki_samples} and fed them into serval such random Transformers with different parameters size. 
In Fig.~\ref{fig:energy_lists_combined_random}, we present the correlator of the outputs from the intermediate layers. 
We can see that the correlator keeps decreasing until it meets a plateau. Once the plateau phase is reached, we can claim that the word vectors are randomly diffused into the $\mathbb{R}^{d_{\text{embed}}}$ space. Therefore, we can use the data from the random Transformer to calculate the constant, this may be related to the fixed point in previous work~\cite{mathattention}:
\begin{figure}[t]
\centerline{\includegraphics[width=0.8\linewidth]{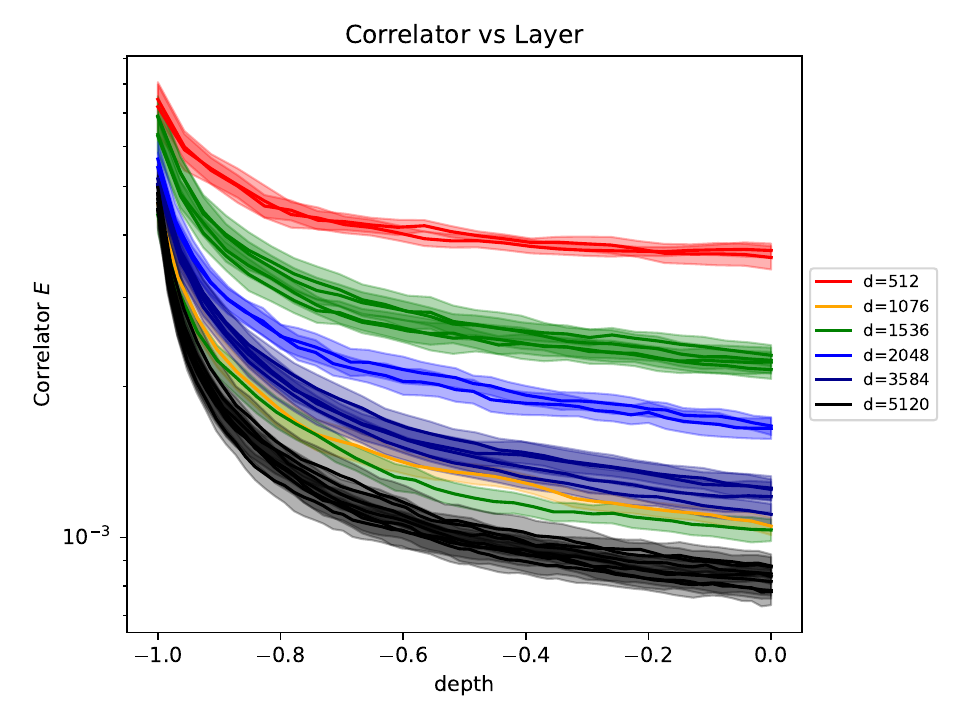}} 
	\caption{How correlator changes in layers of a random model(correlator axis is log scaled)} \label{fig:energy_lists_combined_random}
\end{figure}
\begin{equation}
    \text{Constant} = E_{\text{random}} \cdot (d_{\text{embed}}-1)\,,
\end{equation}
where $E_{\text{random}}$ is the minimum value for correlators  calculated by random Transformer.

With this simple relation, we are able to directly extract $d_{\text{model}}$ and $d_{\text{machine}}$:
\begin{align}
    d_{\text{model}} &= \frac{E_{\text{random}} \cdot (d_{\text{embed}}-1) }{E_{\text{model}}} +1 \nonumber,\quad \\
    d_{\text{machine}} &= \frac{E_{\text{random}} \cdot (d_{\text{embed}}-1) }{E_{\text{machine}}} + 1 \,.
\end{align}

\section{Experiment}
\label{sec:exp}
In this section, we apply the methodology introduced in Section~\ref{section_method} to investigate several models and explicitly calculate the predefined quantities $d_{\text{model}}$ and $d_{\text{machine}}$. We also provide detailed interpretations of these quantities.

To begin, we revisit the evolution of the correlator during the forward propagation. In Fig.~\ref{fig:Qwen2.5-3B}, we calculate the correlator for Sentence Sample 1 using Qwen2.5-3B. This figure illustrates the spectrum of token embeddings for Sample 1 (Appendix~\ref{appendix:wiki_samples}) after processing by Qwen2.5-3B, demonstrating the transition from high-dimensional noise to structured low-dimensional clustering~\cite{layers,layersndtoken,cluster,wordvector}.

\begin{figure}[htbp]
	\centering
	\begin{subfigure}{0.47\linewidth}
		\centering
		\includegraphics[width=1.2\linewidth]{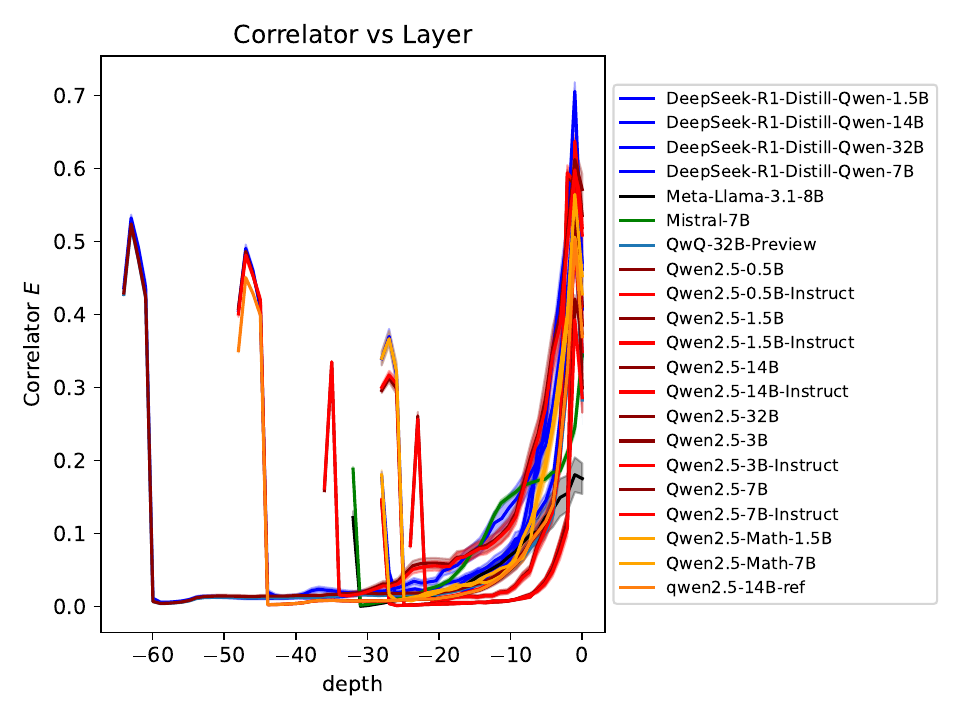}
		\caption{Linear correlator axis}
	\end{subfigure}
	\centering
	\begin{subfigure}{0.47\linewidth}
		\centering
		\includegraphics[width=1.2\linewidth]{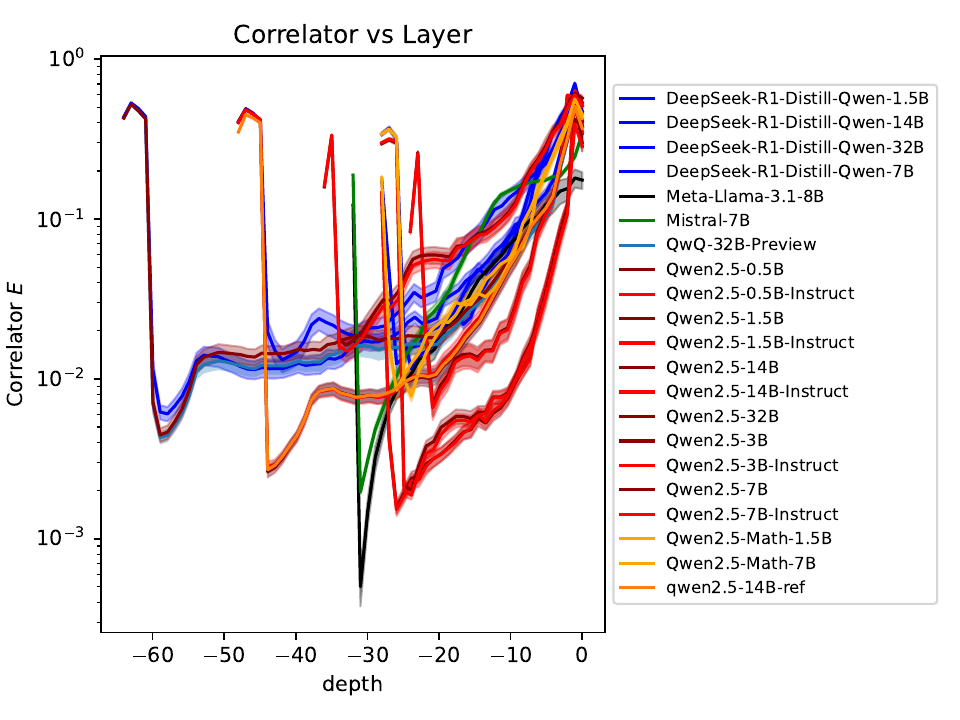}
		\caption{Log correlator axis}
	\end{subfigure}
	\centering
    \caption{Evolution for correlator calculated by other models}
    \label{fig:d_model_models}
\end{figure}

We validate our framework across five distinct text domains (including physics, literature, and biology, etc.) with input lengths ranging from 400 to 1,000 tokens. This ensures robustness to variations in content and context. Additionally, we evaluate multiple models, as shown in Fig.~\ref{fig:d_model_models}. Besides Qwen2.5 series, we include Meta-Llama~\cite{llama3herdmodels}, Deepseek-R1-Distill-Qwen~\cite{deepseekdistill}, and Mistral~\cite{mistral7b}, with varying parameter sizes. In Fig.~\ref{fig:d_model_models}, we plot the correlators for all these models. It is evident that all models exhibit a similar evolution process of the correlator.

Therefore, we consider this to be a general characteristic of LLMs. Specifically, LLMs first diffuse the word vectors into their internal working space \(\mathcal{E}^{*}\), which varies across different models. Within this space, the models initially translate human words into a machine-interpretable feature language. Subsequently, then gradually generate semantic concepts from this machine language. Finally, in the last few layers of the models, the machine language is transformed into a human-like language that the LLMs have learned.

\begin{figure}[t]
\centerline{\includegraphics[width=0.8\linewidth]{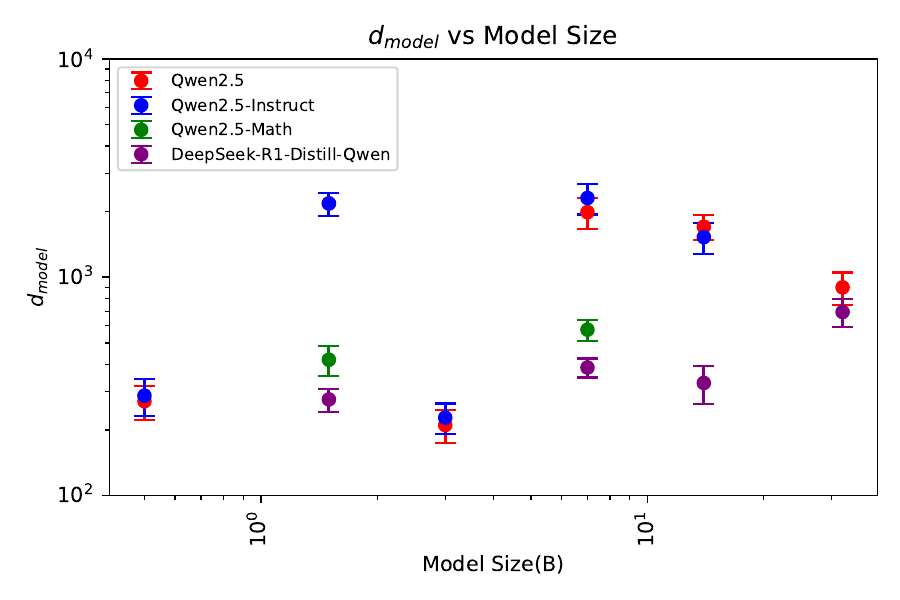}} 
	\caption{ $d_{\text{model}}$ for different model} \label{fig:compare}
\end{figure}

We explicitly calculate \( d_{\text{model}} \) for each model, with the results presented in Fig.~\ref{fig:compare}. The findings reveal that \( d_{\text{model}} \) varies across different models, indicating that each model possesses its unique working space \(\mathcal{E}^*\). For models sharing the same parameter size and architecture, our analysis shows that model performance~\cite{deepseekdistill,Qwenbench,QwenbenchII} is negatively correlated with \( d_{\text{model}} \). This correlation suggests that lower \( d_{\text{model}} \) values imply more efficient compression of task-relevant features, thereby mitigating overfitting to high-dimensional noise.

Although \(d_{\text{model}}\) provides performance insights that are dependent on model weights, its applicability is restricted to models with similar architectures and tokenization schemes. If we wish to conduct cross-paradigm comparisons in the future, careful normalization will be necessary.

We also conjecture that there exists an optimal value for \(d_{\text{model}}\). The observed negative correlation is likely because the \(d_{\text{model}}\) values in our current analysis are all greater than this optimal value.

At the 3B parameter scale, the models in the figure appear peculiar, as their \(d_{\text{model}}\) values are relatively lower than those of other models. We conjecture that this may be due to the extensive optimization that the Qwen2.5-3B model has undergone (especially noting that Qwen2.5-3B has a deeper layer structure with \(L_{\text{max}} = 36\))~\cite{QwenbenchII}.

\begin{figure}[t]
\centerline{\includegraphics[width=0.75\linewidth]{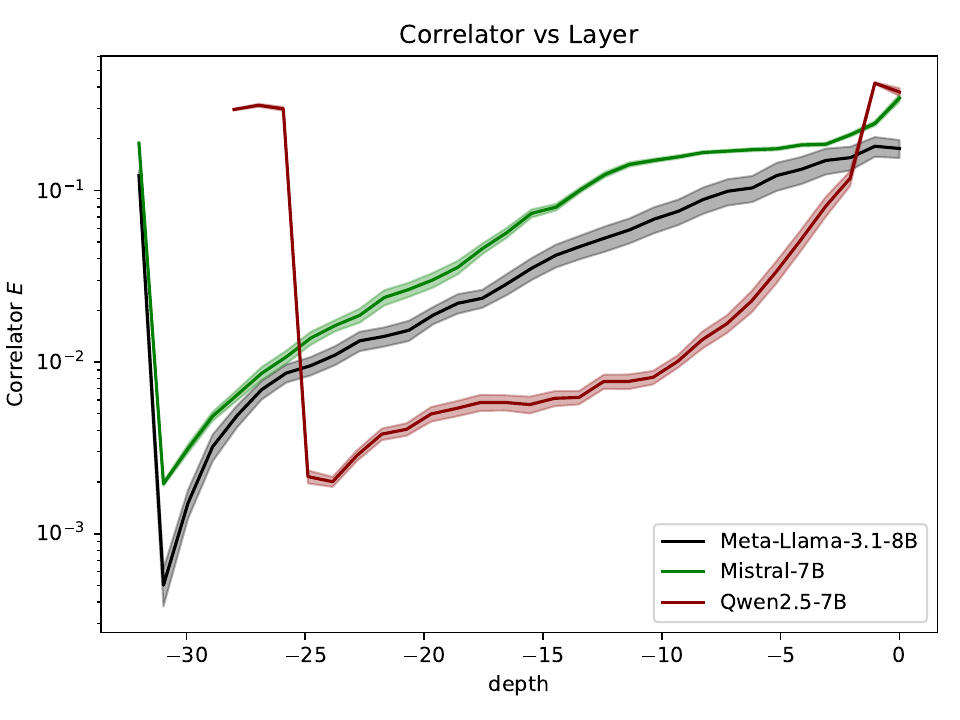}} 
	\caption{ Correlator curves for Qwen2.5-7B, Mistral-7B and Meta-Llama-8B, whose parameters are close yet the shape of curve differs a lot. }\label{fig:energy_lists_combined}
\end{figure}

One might question why we do not compare Qwen2.5-7B and Mistral-7B, despite their identical parameter counts. The reason is that their architectures exhibit significant differences~\cite{mistral7b,llama3herdmodels}, which in turn lead to markedly different shapes of the correlator curves, as illustrated in Fig.~\ref{fig:energy_lists_combined}. Consequently, meaningful comparisons can only be made between models with identical architectures.

\begin{figure}[t]
\centerline{\includegraphics[width=0.8\linewidth]{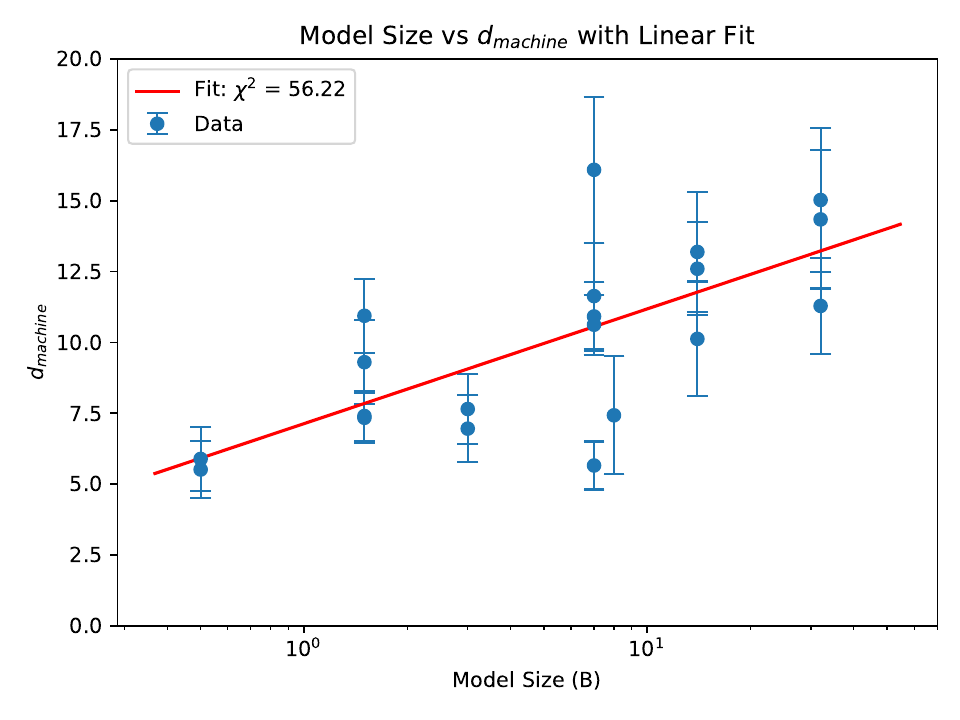}} 
	\caption{ $d_{\text{machine}}$ for different model and linear fitted.} \label{fig:model_size_vs_d_value_with_fit}
\end{figure}

However, it is still useful to compare the dimension \(d_{\text{machine}}\) of these models, as shown in Fig.~\ref{fig:model_size_vs_d_value_with_fit}. To further investigate the properties of \(d_{\text{machine}}\), we calculate this dimension for each model and plot the results in Fig.~\ref{fig:model_size_vs_d_value_with_fit}. 

From the figure, it is evident that \(d_{\text{machine}}\) exhibits a positive linear correlation with the number of parameters, despite some variance between models with the same number of parameters. This trend is reasonable, as a model must reach a certain size to fully understand human language~\cite{idoftext,canbedistinguish,ID}. Beyond this threshold, the emergent phenomena of LLMs enable them to form a semantic manifold \(\mathcal{E}_{\text{machine}}\) that surpasses the human semantic space~\cite{emergewei}.

\begin{figure}
    \centering
    \includegraphics[width=1\linewidth]{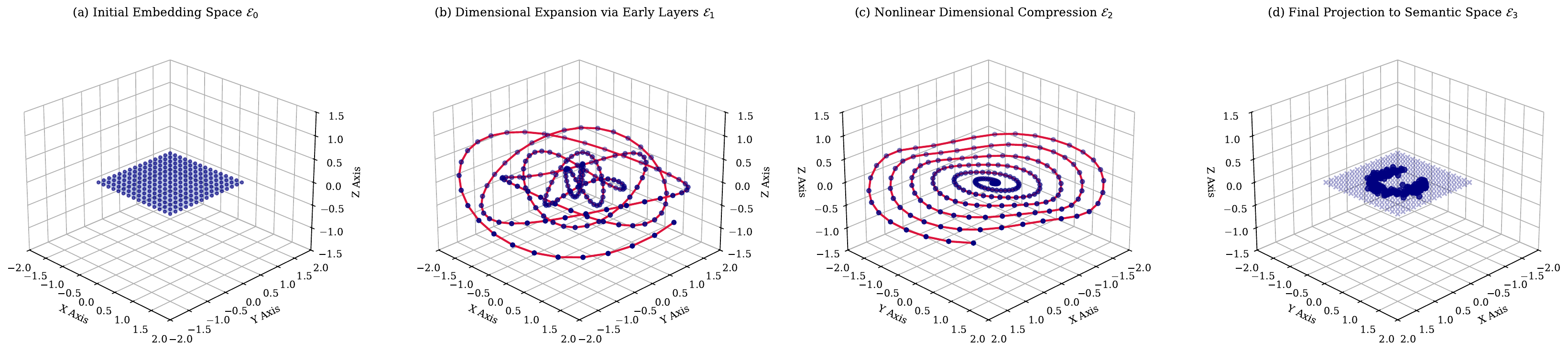}
    \caption{Illustrative representation of the Evolution of Embedding Manifolds in LLMs. The diagram serves as an example with three layers to demonstrate the conceptual development across the model's depth. (a) Initial Embedding Space $\mathcal{E}_0$, where discrete token vectors are embedded on a low-dimensional submanifold. (b) Dimensional Expansion via Early Layers $\mathcal{E}_1$, where the model induces a nonlinear mapping to a high-dimensional working space. (c) Nonlinear Dimensional Compression $\mathcal{E}_2$, where deeper layers compress irrelevant dimensions while preserving task-relevant information. (d) Final Projection to Semantic Space $\mathcal{E}_3$, where the representation is aligned with human-interpretable semantic structures}
    \label{fig:show}
\end{figure}

Through a systematic analysis, we have characterized the operational workflow of Large Language Models (LLMs) as a hierarchical dimensional transformation process. This process can be divided into several distinct phases.

Initially, the model begins with discrete token vectors \( \boldsymbol{t}_i \in \mathbb{R}^{d_{\text{embed}}} \) corresponding to logits. Importantly, these vectors are not uniformly distributed in the high-dimensional embedding space \( \mathbb{R}^{d_{\text{embed}}} \), but rather reside on a low-dimensional submanifold \( \mathcal{E}_0 \subset \mathbb{R}^{d_{\text{embed}}} \) with intrinsic dimension \( \dim(\mathcal{E}_0) < d_{\text{embed}} \). This phenomenon is consistent with the empirical observation that natural language tokens exhibit geometric sparsity in embedding spaces~\cite{idoftext}.

Subsequently, through positional encoding and shallow Transformer layers, the model induces a nonlinear mapping:
\begin{equation}
    \mathcal{E}_0 \xrightarrow{f_{\text{early}}} \mathcal{E}^* \subset \mathbb{R}^{d_{\text{model}}}\,,
\end{equation}
where \( \mathcal{E}^* \) represents the model's high-dimensional working space with \( \dim(\mathcal{E}^*) < d_{\text{embed}} \). This phase generates a spectrally rich representation space, which temporarily sacrifices dimensional efficiency in favor of enhanced computational expressivity~\cite{peakII,layersndtoken,Resnet}.

Next, deeper Transformer layers act as geometric projectors that progressively compress irrelevant dimensions through attention mechanisms and feedforward networks. Formally, this can be modeled as a sequence of smooth mappings:
\begin{equation}
    \mathcal{E}_\xi \xrightarrow{\mathcal{T}_\xi} \mathcal{E}_{\xi+1} \subset \mathbb{R}^{d_{\text{model}}}\,, \quad \dim(\mathcal{E}_{\xi+1}) < \dim(\mathcal{E}_{\xi})\,.
\end{equation}
Each layer \( \xi \) reduces the intrinsic dimension while preserving task-relevant information~\cite{mathattentionIII}.

Finally, in the terminal layers, the compressed representation \( \boldsymbol{t} \in \mathcal{E}_m \) is aligned with the original embedding space via the unembedding transformation:
\begin{equation}
    \boldsymbol{t} \mapsto \text{token id}
    \label{eq:map}\,.
\end{equation}
This geometric alignment ensures that the final prediction resides near human-interpretable semantic structures.

This workflow elucidates two fundamental principles. First, the alignment between \( d_{\text{machine}} \) and human semantic dimensionality~\cite{idoftext} emerges naturally from the model's need to perform the final mapping using Equation~\ref{eq:map}. Second, empirical studies demonstrate a negative relationship between \( d_{\text{model}} \) and model performance. This is attributed to the fact that a lower \( d_{\text{model}} \) indicates that the model achieves a sharper focus on semantically relevant features while suppressing high-dimensional noise~\cite{Qwenbench, deepseekdistill}.

\section{Conclusion}
\label{sec:conclusion}

In this work, we delve into the geometric dynamics of token representations in LLMs by introducing a novel quantity: the correlator $E$. This metric effectively captures the layer-wise evolution patterns and is closely related to intrinsic dimension (ID)~\cite{layers,layersndtoken}. By examining the evolution of the correlator across layers, we can quantitatively characterize the evolution of token dimensions. This approach sets our work apart from previous studies~\cite{layersndtoken,ID}.

Through the lens of correlator dynamics (\(E\)), we demonstrate that modern LLMs mediate between two distinct dimensional regimes: (1) a high-dimensional machine's working space (\(\mathcal{E}^*\), with \(d_{\text{model}} \sim 10^2\)) where features are captured, and (2) a low-dimensional semantic manifold (\(\mathcal{E}_m\), with \(d_{\text{machine}} \sim 10^1\)) that mirrors the organization of human language. This mediation follows a characteristic expansion-contraction pattern. Specifically, tokens initially diffuse into the working space \(\mathcal{E}^*\) through the shallow layers, where features are captured and stored. Subsequently, these features are progressively projected onto the target space \(\mathcal{E}_m\) via deeper transformations. During this process, features are combined, and redundant information is discarded to reconstruct the language.

In particular, we propose several quantities that are noteworthy and potentially useful:

(1) \(\mathcal{E}^*\) and \(d_{\text{model}}\): These represent the "machine's working space" and its dimension, respectively. \(d_{\text{model}}\) may also be negatively correlated with model performance and is of the similar magnitude as that described in previous work~\cite{idofmodel}.

(2) \(\mathcal{E}_m\) and \(d_{\text{machine}}\): These represent the "semantic space learned by machines" and are closely related to the intrinsic dimension of the dataset, as defined in previous work~\cite{idoftext}. They are also analogous to those measured through the outputs of LLMs, as detailed in previous work~\cite{phase}.

As we have seen, these quantities fit within the same framework in our description. Therefore, our work may provide a unifying framework that can integrate the diverse studies on model dimensions from previous research. However, our theory relies on several simplifying assumptions, which may account for the deviations between the \(d_{\text{machine}}\) predicted in this paper and those found in previous research. Establishing a more rigorous mathematical description of our theory that does not rely on these simplifications is an issue that requires further research in the future.

As a practical application, our theory provides a potential method for quickly assessing the capabilities of a model. By simply executing a single forward propagation, one can extract \(d_{\text{model}}\) and use it to compare different models. However, this approach is limited to models with identical architectures and parameter sizes. Nevertheless, it provides an additional metric for evaluating model quality, beyond parameters and architecture. It will also be interesting to incorporate \(d_{\text{model}}\) into the loss function to investigate whether it can aid in training LLMs. We leave this for future work.

\bibliography{cite}

\begin{thebibliography}{38}
\expandafter\ifx\csname natexlab\endcsname\relax\def\natexlab#1{#1}\fi
\expandafter\ifx\csname bibnamefont\endcsname\relax
  \def\bibnamefont#1{#1}\fi
\expandafter\ifx\csname bibfnamefont\endcsname\relax
  \def\bibfnamefont#1{#1}\fi
\expandafter\ifx\csname citenamefont\endcsname\relax
  \def\citenamefont#1{#1}\fi
\expandafter\ifx\csname url\endcsname\relax
  \def\url#1{\texttt{#1}}\fi
\expandafter\ifx\csname urlprefix\endcsname\relax\def\urlprefix{URL }\fi
\providecommand{\bibinfo}[2]{#2}
\providecommand{\eprint}[2][]{\url{#2}}

\bibitem[{\citenamefont{Tulchinskii et~al.}(2023)\citenamefont{Tulchinskii, Kuznetsov, Kushnareva, Cherniavskii, Barannikov, Piontkovskaya, Nikolenko, and Burnaev}}]{idoftext}
\bibinfo{author}{\bibfnamefont{E.}~\bibnamefont{Tulchinskii}}, \bibinfo{author}{\bibfnamefont{K.}~\bibnamefont{Kuznetsov}}, \bibinfo{author}{\bibfnamefont{L.}~\bibnamefont{Kushnareva}}, \bibinfo{author}{\bibfnamefont{D.}~\bibnamefont{Cherniavskii}}, \bibinfo{author}{\bibfnamefont{S.}~\bibnamefont{Barannikov}}, \bibinfo{author}{\bibfnamefont{I.}~\bibnamefont{Piontkovskaya}}, \bibinfo{author}{\bibfnamefont{S.}~\bibnamefont{Nikolenko}}, \bibnamefont{and} \bibinfo{author}{\bibfnamefont{E.}~\bibnamefont{Burnaev}}, \emph{\bibinfo{title}{Intrinsic dimension estimation for robust detection of ai-generated texts}} (\bibinfo{year}{2023}), \eprint{2306.04723}, \urlprefix\url{https://arxiv.org/abs/2306.04723}.

\bibitem[{\citenamefont{Mikolov et~al.}(2013)\citenamefont{Mikolov, Chen, Corrado, and Dean}}]{wordvector}
\bibinfo{author}{\bibfnamefont{T.}~\bibnamefont{Mikolov}}, \bibinfo{author}{\bibfnamefont{K.}~\bibnamefont{Chen}}, \bibinfo{author}{\bibfnamefont{G.}~\bibnamefont{Corrado}}, \bibnamefont{and} \bibinfo{author}{\bibfnamefont{J.}~\bibnamefont{Dean}}, \emph{\bibinfo{title}{Efficient estimation of word representations in vector space}} (\bibinfo{year}{2013}), \eprint{1301.3781}, \urlprefix\url{https://arxiv.org/abs/1301.3781}.

\bibitem[{\citenamefont{Vaswani et~al.}(2023)\citenamefont{Vaswani, Shazeer, Parmar, Uszkoreit, Jones, Gomez, Kaiser, and Polosukhin}}]{attentionneed}
\bibinfo{author}{\bibfnamefont{A.}~\bibnamefont{Vaswani}}, \bibinfo{author}{\bibfnamefont{N.}~\bibnamefont{Shazeer}}, \bibinfo{author}{\bibfnamefont{N.}~\bibnamefont{Parmar}}, \bibinfo{author}{\bibfnamefont{J.}~\bibnamefont{Uszkoreit}}, \bibinfo{author}{\bibfnamefont{L.}~\bibnamefont{Jones}}, \bibinfo{author}{\bibfnamefont{A.~N.} \bibnamefont{Gomez}}, \bibinfo{author}{\bibfnamefont{L.}~\bibnamefont{Kaiser}}, \bibnamefont{and} \bibinfo{author}{\bibfnamefont{I.}~\bibnamefont{Polosukhin}}, \emph{\bibinfo{title}{Attention is all you need}} (\bibinfo{year}{2023}), \eprint{1706.03762}, \urlprefix\url{https://arxiv.org/abs/1706.03762}.

\bibitem[{\citenamefont{Vuckovic et~al.}(2020)\citenamefont{Vuckovic, Baratin, and des Combes}}]{mathattention}
\bibinfo{author}{\bibfnamefont{J.}~\bibnamefont{Vuckovic}}, \bibinfo{author}{\bibfnamefont{A.}~\bibnamefont{Baratin}}, \bibnamefont{and} \bibinfo{author}{\bibfnamefont{R.~T.} \bibnamefont{des Combes}}, \emph{\bibinfo{title}{A mathematical theory of attention}} (\bibinfo{year}{2020}), \eprint{2007.02876}, \urlprefix\url{https://arxiv.org/abs/2007.02876}.

\bibitem[{\citenamefont{Geshkovski et~al.}(2024{\natexlab{a}})\citenamefont{Geshkovski, Letrouit, Polyanskiy, and Rigollet}}]{mathattentionII}
\bibinfo{author}{\bibfnamefont{B.}~\bibnamefont{Geshkovski}}, \bibinfo{author}{\bibfnamefont{C.}~\bibnamefont{Letrouit}}, \bibinfo{author}{\bibfnamefont{Y.}~\bibnamefont{Polyanskiy}}, \bibnamefont{and} \bibinfo{author}{\bibfnamefont{P.}~\bibnamefont{Rigollet}}, \emph{\bibinfo{title}{A mathematical perspective on transformers}} (\bibinfo{year}{2024}{\natexlab{a}}), \eprint{2312.10794}, \urlprefix\url{https://arxiv.org/abs/2312.10794}.

\bibitem[{\citenamefont{Cowsik et~al.}(2024)\citenamefont{Cowsik, Nebabu, Qi, and Ganguli}}]{mathattentionIII}
\bibinfo{author}{\bibfnamefont{A.}~\bibnamefont{Cowsik}}, \bibinfo{author}{\bibfnamefont{T.}~\bibnamefont{Nebabu}}, \bibinfo{author}{\bibfnamefont{X.-L.} \bibnamefont{Qi}}, \bibnamefont{and} \bibinfo{author}{\bibfnamefont{S.}~\bibnamefont{Ganguli}}, \emph{\bibinfo{title}{Geometric dynamics of signal propagation predict trainability of transformers}} (\bibinfo{year}{2024}), \eprint{2403.02579}, \urlprefix\url{https://arxiv.org/abs/2403.02579}.

\bibitem[{\citenamefont{Sun and Haghighat}(2025)}]{phase}
\bibinfo{author}{\bibfnamefont{Y.}~\bibnamefont{Sun}} \bibnamefont{and} \bibinfo{author}{\bibfnamefont{B.}~\bibnamefont{Haghighat}}, \emph{\bibinfo{title}{Phase transitions in large language models and the $o(n)$ model}} (\bibinfo{year}{2025}), \eprint{2501.16241}, \urlprefix\url{https://arxiv.org/abs/2501.16241}.

\bibitem[{\citenamefont{Li et~al.}(2018)\citenamefont{Li, Farkhoor, Liu, and Yosinski}}]{dimensionstart}
\bibinfo{author}{\bibfnamefont{C.}~\bibnamefont{Li}}, \bibinfo{author}{\bibfnamefont{H.}~\bibnamefont{Farkhoor}}, \bibinfo{author}{\bibfnamefont{R.}~\bibnamefont{Liu}}, \bibnamefont{and} \bibinfo{author}{\bibfnamefont{J.}~\bibnamefont{Yosinski}}, \emph{\bibinfo{title}{Measuring the intrinsic dimension of objective landscapes}} (\bibinfo{year}{2018}), \eprint{1804.08838}, \urlprefix\url{https://arxiv.org/abs/1804.08838}.

\bibitem[{\citenamefont{Ansuini et~al.}(2019)\citenamefont{Ansuini, Laio, Macke, and Zoccolan}}]{ID}
\bibinfo{author}{\bibfnamefont{A.}~\bibnamefont{Ansuini}}, \bibinfo{author}{\bibfnamefont{A.}~\bibnamefont{Laio}}, \bibinfo{author}{\bibfnamefont{J.~H.} \bibnamefont{Macke}}, \bibnamefont{and} \bibinfo{author}{\bibfnamefont{D.}~\bibnamefont{Zoccolan}}, \emph{\bibinfo{title}{Intrinsic dimension of data representations in deep neural networks}} (\bibinfo{year}{2019}), \eprint{1905.12784}, \urlprefix\url{https://arxiv.org/abs/1905.12784}.

\bibitem[{\citenamefont{Cheng et~al.}(2023)\citenamefont{Cheng, Kervadec, and Baroni}}]{IDhavemeaning}
\bibinfo{author}{\bibfnamefont{E.}~\bibnamefont{Cheng}}, \bibinfo{author}{\bibfnamefont{C.}~\bibnamefont{Kervadec}}, \bibnamefont{and} \bibinfo{author}{\bibfnamefont{M.}~\bibnamefont{Baroni}}, \emph{\bibinfo{title}{Bridging information-theoretic and geometric compression in language models}} (\bibinfo{year}{2023}), \eprint{2310.13620}, \urlprefix\url{https://arxiv.org/abs/2310.13620}.

\bibitem[{\citenamefont{Pope et~al.}(2021{\natexlab{a}})\citenamefont{Pope, Zhu, Abdelkader, Goldblum, and Goldstein}}]{IDofimage}
\bibinfo{author}{\bibfnamefont{P.}~\bibnamefont{Pope}}, \bibinfo{author}{\bibfnamefont{C.}~\bibnamefont{Zhu}}, \bibinfo{author}{\bibfnamefont{A.}~\bibnamefont{Abdelkader}}, \bibinfo{author}{\bibfnamefont{M.}~\bibnamefont{Goldblum}}, \bibnamefont{and} \bibinfo{author}{\bibfnamefont{T.}~\bibnamefont{Goldstein}}, \emph{\bibinfo{title}{The intrinsic dimension of images and its impact on learning}} (\bibinfo{year}{2021}{\natexlab{a}}), \eprint{2104.08894}, \urlprefix\url{https://arxiv.org/abs/2104.08894}.

\bibitem[{\citenamefont{Aghajanyan et~al.}(2020)\citenamefont{Aghajanyan, Zettlemoyer, and Gupta}}]{idofmodel}
\bibinfo{author}{\bibfnamefont{A.}~\bibnamefont{Aghajanyan}}, \bibinfo{author}{\bibfnamefont{L.}~\bibnamefont{Zettlemoyer}}, \bibnamefont{and} \bibinfo{author}{\bibfnamefont{S.}~\bibnamefont{Gupta}}, \emph{\bibinfo{title}{Intrinsic dimensionality explains the effectiveness of language model fine-tuning}} (\bibinfo{year}{2020}), \eprint{2012.13255}, \urlprefix\url{https://arxiv.org/abs/2012.13255}.

\bibitem[{\citenamefont{Pope et~al.}(2021{\natexlab{b}})\citenamefont{Pope, Zhu, Abdelkader, Goldblum, and Goldstein}}]{imagedimension}
\bibinfo{author}{\bibfnamefont{P.}~\bibnamefont{Pope}}, \bibinfo{author}{\bibfnamefont{C.}~\bibnamefont{Zhu}}, \bibinfo{author}{\bibfnamefont{A.}~\bibnamefont{Abdelkader}}, \bibinfo{author}{\bibfnamefont{M.}~\bibnamefont{Goldblum}}, \bibnamefont{and} \bibinfo{author}{\bibfnamefont{T.}~\bibnamefont{Goldstein}}, \emph{\bibinfo{title}{The intrinsic dimension of images and its impact on learning}} (\bibinfo{year}{2021}{\natexlab{b}}), \eprint{2104.08894}, \urlprefix\url{https://arxiv.org/abs/2104.08894}.

\bibitem[{\citenamefont{Valeriani et~al.}(2023)\citenamefont{Valeriani, Doimo, Cuturello, Laio, Ansuini, and Cazzaniga}}]{layers}
\bibinfo{author}{\bibfnamefont{L.}~\bibnamefont{Valeriani}}, \bibinfo{author}{\bibfnamefont{D.}~\bibnamefont{Doimo}}, \bibinfo{author}{\bibfnamefont{F.}~\bibnamefont{Cuturello}}, \bibinfo{author}{\bibfnamefont{A.}~\bibnamefont{Laio}}, \bibinfo{author}{\bibfnamefont{A.}~\bibnamefont{Ansuini}}, \bibnamefont{and} \bibinfo{author}{\bibfnamefont{A.}~\bibnamefont{Cazzaniga}}, \emph{\bibinfo{title}{The geometry of hidden representations of large transformer models}} (\bibinfo{year}{2023}), \eprint{2302.00294}, \urlprefix\url{https://arxiv.org/abs/2302.00294}.

\bibitem[{\citenamefont{Viswanathan et~al.}(2025)\citenamefont{Viswanathan, Gardinazzi, Panerai, Cazzaniga, and Biagetti}}]{layersndtoken}
\bibinfo{author}{\bibfnamefont{K.}~\bibnamefont{Viswanathan}}, \bibinfo{author}{\bibfnamefont{Y.}~\bibnamefont{Gardinazzi}}, \bibinfo{author}{\bibfnamefont{G.}~\bibnamefont{Panerai}}, \bibinfo{author}{\bibfnamefont{A.}~\bibnamefont{Cazzaniga}}, \bibnamefont{and} \bibinfo{author}{\bibfnamefont{M.}~\bibnamefont{Biagetti}}, \emph{\bibinfo{title}{The geometry of tokens in internal representations of large language models}} (\bibinfo{year}{2025}), \eprint{2501.10573}, \urlprefix\url{https://arxiv.org/abs/2501.10573}.

\bibitem[{\citenamefont{Yang et~al.}(2024)\citenamefont{Yang, Zhang, Hui, Gao, Yu, Li, Liu, Tu, Zhou, Lin et~al.}}]{Qwenbench}
\bibinfo{author}{\bibfnamefont{A.}~\bibnamefont{Yang}}, \bibinfo{author}{\bibfnamefont{B.}~\bibnamefont{Zhang}}, \bibinfo{author}{\bibfnamefont{B.}~\bibnamefont{Hui}}, \bibinfo{author}{\bibfnamefont{B.}~\bibnamefont{Gao}}, \bibinfo{author}{\bibfnamefont{B.}~\bibnamefont{Yu}}, \bibinfo{author}{\bibfnamefont{C.}~\bibnamefont{Li}}, \bibinfo{author}{\bibfnamefont{D.}~\bibnamefont{Liu}}, \bibinfo{author}{\bibfnamefont{J.}~\bibnamefont{Tu}}, \bibinfo{author}{\bibfnamefont{J.}~\bibnamefont{Zhou}}, \bibinfo{author}{\bibfnamefont{J.}~\bibnamefont{Lin}}, \bibnamefont{et~al.}, \emph{\bibinfo{title}{Qwen2.5-math technical report: Toward mathematical expert model via self-improvement}} (\bibinfo{year}{2024}), \eprint{2409.12122}, \urlprefix\url{https://arxiv.org/abs/2409.12122}.

\bibitem[{\citenamefont{Qwen et~al.}(2025)\citenamefont{Qwen, :, Yang, Yang, Zhang, Hui, Zheng, Yu, Li, Liu et~al.}}]{QwenbenchII}
\bibinfo{author}{\bibnamefont{Qwen}}, \bibinfo{author}{\bibnamefont{:}}, \bibinfo{author}{\bibfnamefont{A.}~\bibnamefont{Yang}}, \bibinfo{author}{\bibfnamefont{B.}~\bibnamefont{Yang}}, \bibinfo{author}{\bibfnamefont{B.}~\bibnamefont{Zhang}}, \bibinfo{author}{\bibfnamefont{B.}~\bibnamefont{Hui}}, \bibinfo{author}{\bibfnamefont{B.}~\bibnamefont{Zheng}}, \bibinfo{author}{\bibfnamefont{B.}~\bibnamefont{Yu}}, \bibinfo{author}{\bibfnamefont{C.}~\bibnamefont{Li}}, \bibinfo{author}{\bibfnamefont{D.}~\bibnamefont{Liu}}, \bibnamefont{et~al.}, \emph{\bibinfo{title}{Qwen2.5 technical report}} (\bibinfo{year}{2025}), \eprint{2412.15115}, \urlprefix\url{https://arxiv.org/abs/2412.15115}.

\bibitem[{\citenamefont{DeepSeek-AI}(2025)}]{deepseekdistill}
\bibinfo{author}{\bibnamefont{DeepSeek-AI}}, \emph{\bibinfo{title}{Deepseek-r1: Incentivizing reasoning capability in llms via reinforcement learning}} (\bibinfo{year}{2025}), \eprint{2501.12948}, \urlprefix\url{https://arxiv.org/abs/2501.12948}.

\bibitem[{\citenamefont{Belrose et~al.}(2023)\citenamefont{Belrose, Furman, Smith, Halawi, Ostrovsky, McKinney, Biderman, and Steinhardt}}]{wordtrack}
\bibinfo{author}{\bibfnamefont{N.}~\bibnamefont{Belrose}}, \bibinfo{author}{\bibfnamefont{Z.}~\bibnamefont{Furman}}, \bibinfo{author}{\bibfnamefont{L.}~\bibnamefont{Smith}}, \bibinfo{author}{\bibfnamefont{D.}~\bibnamefont{Halawi}}, \bibinfo{author}{\bibfnamefont{I.}~\bibnamefont{Ostrovsky}}, \bibinfo{author}{\bibfnamefont{L.}~\bibnamefont{McKinney}}, \bibinfo{author}{\bibfnamefont{S.}~\bibnamefont{Biderman}}, \bibnamefont{and} \bibinfo{author}{\bibfnamefont{J.}~\bibnamefont{Steinhardt}}, \emph{\bibinfo{title}{Eliciting latent predictions from transformers with the tuned lens}} (\bibinfo{year}{2023}), \eprint{2303.08112}, \urlprefix\url{https://arxiv.org/abs/2303.08112}.

\bibitem[{\citenamefont{Doimo et~al.}(2020)\citenamefont{Doimo, Glielmo, Ansuini, and Laio}}]{peakII}
\bibinfo{author}{\bibfnamefont{D.}~\bibnamefont{Doimo}}, \bibinfo{author}{\bibfnamefont{A.}~\bibnamefont{Glielmo}}, \bibinfo{author}{\bibfnamefont{A.}~\bibnamefont{Ansuini}}, \bibnamefont{and} \bibinfo{author}{\bibfnamefont{A.}~\bibnamefont{Laio}}, \emph{\bibinfo{title}{Hierarchical nucleation in deep neural networks}} (\bibinfo{year}{2020}), \eprint{2007.03506}, \urlprefix\url{https://arxiv.org/abs/2007.03506}.

\bibitem[{\citenamefont{Jiang et~al.}(2023)\citenamefont{Jiang, Sablayrolles, Mensch, Bamford, Chaplot, de~las Casas, Bressand, Lengyel, Lample, Saulnier et~al.}}]{mistral7b}
\bibinfo{author}{\bibfnamefont{A.~Q.} \bibnamefont{Jiang}}, \bibinfo{author}{\bibfnamefont{A.}~\bibnamefont{Sablayrolles}}, \bibinfo{author}{\bibfnamefont{A.}~\bibnamefont{Mensch}}, \bibinfo{author}{\bibfnamefont{C.}~\bibnamefont{Bamford}}, \bibinfo{author}{\bibfnamefont{D.~S.} \bibnamefont{Chaplot}}, \bibinfo{author}{\bibfnamefont{D.}~\bibnamefont{de~las Casas}}, \bibinfo{author}{\bibfnamefont{F.}~\bibnamefont{Bressand}}, \bibinfo{author}{\bibfnamefont{G.}~\bibnamefont{Lengyel}}, \bibinfo{author}{\bibfnamefont{G.}~\bibnamefont{Lample}}, \bibinfo{author}{\bibfnamefont{L.}~\bibnamefont{Saulnier}}, \bibnamefont{et~al.}, \emph{\bibinfo{title}{Mistral 7b}} (\bibinfo{year}{2023}), \eprint{2310.06825}, \urlprefix\url{https://arxiv.org/abs/2310.06825}.

\bibitem[{\citenamefont{Grattafiori et~al.}(2024)\citenamefont{Grattafiori, Dubey, Jauhri, Pandey, Kadian, Al-Dahle, Letman, Mathur, Schelten, Vaughan et~al.}}]{llama3herdmodels}
\bibinfo{author}{\bibfnamefont{A.}~\bibnamefont{Grattafiori}}, \bibinfo{author}{\bibfnamefont{A.}~\bibnamefont{Dubey}}, \bibinfo{author}{\bibfnamefont{A.}~\bibnamefont{Jauhri}}, \bibinfo{author}{\bibfnamefont{A.}~\bibnamefont{Pandey}}, \bibinfo{author}{\bibfnamefont{A.}~\bibnamefont{Kadian}}, \bibinfo{author}{\bibfnamefont{A.}~\bibnamefont{Al-Dahle}}, \bibinfo{author}{\bibfnamefont{A.}~\bibnamefont{Letman}}, \bibinfo{author}{\bibfnamefont{A.}~\bibnamefont{Mathur}}, \bibinfo{author}{\bibfnamefont{A.}~\bibnamefont{Schelten}}, \bibinfo{author}{\bibfnamefont{A.}~\bibnamefont{Vaughan}}, \bibnamefont{et~al.}, \emph{\bibinfo{title}{The llama 3 herd of models}} (\bibinfo{year}{2024}), \eprint{2407.21783}, \urlprefix\url{https://arxiv.org/abs/2407.21783}.

\bibitem[{\citenamefont{Kushnareva et~al.}(2021)\citenamefont{Kushnareva, Cherniavskii, Mikhailov, Artemova, Barannikov, Bernstein, Piontkovskaya, Piontkovski, and Burnaev}}]{attentiondimension}
\bibinfo{author}{\bibfnamefont{L.}~\bibnamefont{Kushnareva}}, \bibinfo{author}{\bibfnamefont{D.}~\bibnamefont{Cherniavskii}}, \bibinfo{author}{\bibfnamefont{V.}~\bibnamefont{Mikhailov}}, \bibinfo{author}{\bibfnamefont{E.}~\bibnamefont{Artemova}}, \bibinfo{author}{\bibfnamefont{S.}~\bibnamefont{Barannikov}}, \bibinfo{author}{\bibfnamefont{A.}~\bibnamefont{Bernstein}}, \bibinfo{author}{\bibfnamefont{I.}~\bibnamefont{Piontkovskaya}}, \bibinfo{author}{\bibfnamefont{D.}~\bibnamefont{Piontkovski}}, \bibnamefont{and} \bibinfo{author}{\bibfnamefont{E.}~\bibnamefont{Burnaev}}, in \emph{\bibinfo{booktitle}{Proceedings of the 2021 Conference on Empirical Methods in Natural Language Processing}} (\bibinfo{publisher}{Association for Computational Linguistics}, \bibinfo{year}{2021}), p. \bibinfo{pages}{635–649}, \urlprefix\url{http://dx.doi.org/10.18653/v1/2021.emnlp-main.50}.

\bibitem[{\citenamefont{Barannikov et~al.}(2021)\citenamefont{Barannikov, Trofimov, Sotnikov, Trimbach, Korotin, Filippov, and Burnaev}}]{topologicalframe}
\bibinfo{author}{\bibfnamefont{S.}~\bibnamefont{Barannikov}}, \bibinfo{author}{\bibfnamefont{I.}~\bibnamefont{Trofimov}}, \bibinfo{author}{\bibfnamefont{G.}~\bibnamefont{Sotnikov}}, \bibinfo{author}{\bibfnamefont{E.}~\bibnamefont{Trimbach}}, \bibinfo{author}{\bibfnamefont{A.}~\bibnamefont{Korotin}}, \bibinfo{author}{\bibfnamefont{A.}~\bibnamefont{Filippov}}, \bibnamefont{and} \bibinfo{author}{\bibfnamefont{E.}~\bibnamefont{Burnaev}}, \emph{\bibinfo{title}{Manifold topology divergence: a framework for comparing data manifolds}} (\bibinfo{year}{2021}), \eprint{2106.04024}, \urlprefix\url{https://arxiv.org/abs/2106.04024}.

\bibitem[{\citenamefont{Mitchell et~al.}(2023)\citenamefont{Mitchell, Lee, Khazatsky, Manning, and Finn}}]{canbedistinguish}
\bibinfo{author}{\bibfnamefont{E.}~\bibnamefont{Mitchell}}, \bibinfo{author}{\bibfnamefont{Y.}~\bibnamefont{Lee}}, \bibinfo{author}{\bibfnamefont{A.}~\bibnamefont{Khazatsky}}, \bibinfo{author}{\bibfnamefont{C.~D.} \bibnamefont{Manning}}, \bibnamefont{and} \bibinfo{author}{\bibfnamefont{C.}~\bibnamefont{Finn}}, \emph{\bibinfo{title}{Detectgpt: Zero-shot machine-generated text detection using probability curvature}} (\bibinfo{year}{2023}), \eprint{2301.11305}, \urlprefix\url{https://arxiv.org/abs/2301.11305}.

\bibitem[{\citenamefont{Noci et~al.}(2022)\citenamefont{Noci, Anagnostidis, Biggio, Orvieto, Singh, and Lucchi}}]{rank}
\bibinfo{author}{\bibfnamefont{L.}~\bibnamefont{Noci}}, \bibinfo{author}{\bibfnamefont{S.}~\bibnamefont{Anagnostidis}}, \bibinfo{author}{\bibfnamefont{L.}~\bibnamefont{Biggio}}, \bibinfo{author}{\bibfnamefont{A.}~\bibnamefont{Orvieto}}, \bibinfo{author}{\bibfnamefont{S.~P.} \bibnamefont{Singh}}, \bibnamefont{and} \bibinfo{author}{\bibfnamefont{A.}~\bibnamefont{Lucchi}}, \emph{\bibinfo{title}{Signal propagation in transformers: Theoretical perspectives and the role of rank collapse}} (\bibinfo{year}{2022}), \eprint{2206.03126}, \urlprefix\url{https://arxiv.org/abs/2206.03126}.

\bibitem[{\citenamefont{Lan et~al.}(2020)\citenamefont{Lan, Chen, Goodman, Gimpel, Sharma, and Soricut}}]{ALBERT}
\bibinfo{author}{\bibfnamefont{Z.}~\bibnamefont{Lan}}, \bibinfo{author}{\bibfnamefont{M.}~\bibnamefont{Chen}}, \bibinfo{author}{\bibfnamefont{S.}~\bibnamefont{Goodman}}, \bibinfo{author}{\bibfnamefont{K.}~\bibnamefont{Gimpel}}, \bibinfo{author}{\bibfnamefont{P.}~\bibnamefont{Sharma}}, \bibnamefont{and} \bibinfo{author}{\bibfnamefont{R.}~\bibnamefont{Soricut}}, \emph{\bibinfo{title}{Albert: A lite bert for self-supervised learning of language representations}} (\bibinfo{year}{2020}), \eprint{1909.11942}, \urlprefix\url{https://arxiv.org/abs/1909.11942}.

\bibitem[{\citenamefont{He et~al.}(2023)\citenamefont{He, Martens, Zhang, Botev, Brock, Smith, and Teh}}]{differentarc}
\bibinfo{author}{\bibfnamefont{B.}~\bibnamefont{He}}, \bibinfo{author}{\bibfnamefont{J.}~\bibnamefont{Martens}}, \bibinfo{author}{\bibfnamefont{G.}~\bibnamefont{Zhang}}, \bibinfo{author}{\bibfnamefont{A.}~\bibnamefont{Botev}}, \bibinfo{author}{\bibfnamefont{A.}~\bibnamefont{Brock}}, \bibinfo{author}{\bibfnamefont{S.~L.} \bibnamefont{Smith}}, \bibnamefont{and} \bibinfo{author}{\bibfnamefont{Y.~W.} \bibnamefont{Teh}}, \emph{\bibinfo{title}{Deep transformers without shortcuts: Modifying self-attention for faithful signal propagation}} (\bibinfo{year}{2023}), \eprint{2302.10322}, \urlprefix\url{https://arxiv.org/abs/2302.10322}.

\bibitem[{\citenamefont{Geshkovski et~al.}(2024{\natexlab{b}})\citenamefont{Geshkovski, Letrouit, Polyanskiy, and Rigollet}}]{cluster}
\bibinfo{author}{\bibfnamefont{B.}~\bibnamefont{Geshkovski}}, \bibinfo{author}{\bibfnamefont{C.}~\bibnamefont{Letrouit}}, \bibinfo{author}{\bibfnamefont{Y.}~\bibnamefont{Polyanskiy}}, \bibnamefont{and} \bibinfo{author}{\bibfnamefont{P.}~\bibnamefont{Rigollet}}, \emph{\bibinfo{title}{The emergence of clusters in self-attention dynamics}} (\bibinfo{year}{2024}{\natexlab{b}}), \eprint{2305.05465}, \urlprefix\url{https://arxiv.org/abs/2305.05465}.

\bibitem[{\citenamefont{Jastrzębski et~al.}(2018)\citenamefont{Jastrzębski, Arpit, Ballas, Verma, Che, and Bengio}}]{Resnet}
\bibinfo{author}{\bibfnamefont{S.}~\bibnamefont{Jastrzębski}}, \bibinfo{author}{\bibfnamefont{D.}~\bibnamefont{Arpit}}, \bibinfo{author}{\bibfnamefont{N.}~\bibnamefont{Ballas}}, \bibinfo{author}{\bibfnamefont{V.}~\bibnamefont{Verma}}, \bibinfo{author}{\bibfnamefont{T.}~\bibnamefont{Che}}, \bibnamefont{and} \bibinfo{author}{\bibfnamefont{Y.}~\bibnamefont{Bengio}}, \emph{\bibinfo{title}{Residual connections encourage iterative inference}} (\bibinfo{year}{2018}), \eprint{1710.04773}, \urlprefix\url{https://arxiv.org/abs/1710.04773}.

\bibitem[{\citenamefont{Shi et~al.}(2022)\citenamefont{Shi, Gao, Xu, Liang, Li, Kong, Lee, and Kwok}}]{oversmooth}
\bibinfo{author}{\bibfnamefont{H.}~\bibnamefont{Shi}}, \bibinfo{author}{\bibfnamefont{J.}~\bibnamefont{Gao}}, \bibinfo{author}{\bibfnamefont{H.}~\bibnamefont{Xu}}, \bibinfo{author}{\bibfnamefont{X.}~\bibnamefont{Liang}}, \bibinfo{author}{\bibfnamefont{Z.}~\bibnamefont{Li}}, \bibinfo{author}{\bibfnamefont{L.}~\bibnamefont{Kong}}, \bibinfo{author}{\bibfnamefont{S.~M.~S.} \bibnamefont{Lee}}, \bibnamefont{and} \bibinfo{author}{\bibfnamefont{J.~T.} \bibnamefont{Kwok}}, \emph{\bibinfo{title}{Revisiting over-smoothing in bert from the perspective of graph}} (\bibinfo{year}{2022}), \eprint{2202.08625}, \urlprefix\url{https://arxiv.org/abs/2202.08625}.

\bibitem[{\citenamefont{Wu et~al.}(2024)\citenamefont{Wu, Ajorlou, Wu, and Jadbabaie}}]{oversmoothII}
\bibinfo{author}{\bibfnamefont{X.}~\bibnamefont{Wu}}, \bibinfo{author}{\bibfnamefont{A.}~\bibnamefont{Ajorlou}}, \bibinfo{author}{\bibfnamefont{Z.}~\bibnamefont{Wu}}, \bibnamefont{and} \bibinfo{author}{\bibfnamefont{A.}~\bibnamefont{Jadbabaie}}, \emph{\bibinfo{title}{Demystifying oversmoothing in attention-based graph neural networks}} (\bibinfo{year}{2024}), \eprint{2305.16102}, \urlprefix\url{https://arxiv.org/abs/2305.16102}.

\bibitem[{\citenamefont{Wei et~al.}(2022)\citenamefont{Wei, Tay, Bommasani, Raffel, Zoph, Borgeaud, Yogatama, Bosma, Zhou, Metzler et~al.}}]{emergewei}
\bibinfo{author}{\bibfnamefont{J.}~\bibnamefont{Wei}}, \bibinfo{author}{\bibfnamefont{Y.}~\bibnamefont{Tay}}, \bibinfo{author}{\bibfnamefont{R.}~\bibnamefont{Bommasani}}, \bibinfo{author}{\bibfnamefont{C.}~\bibnamefont{Raffel}}, \bibinfo{author}{\bibfnamefont{B.}~\bibnamefont{Zoph}}, \bibinfo{author}{\bibfnamefont{S.}~\bibnamefont{Borgeaud}}, \bibinfo{author}{\bibfnamefont{D.}~\bibnamefont{Yogatama}}, \bibinfo{author}{\bibfnamefont{M.}~\bibnamefont{Bosma}}, \bibinfo{author}{\bibfnamefont{D.}~\bibnamefont{Zhou}}, \bibinfo{author}{\bibfnamefont{D.}~\bibnamefont{Metzler}}, \bibnamefont{et~al.}, \emph{\bibinfo{title}{Emergent abilities of large language models}} (\bibinfo{year}{2022}), \eprint{2206.07682}, \urlprefix\url{https://arxiv.org/abs/2206.07682}.

\bibitem[{\citenamefont{Wikipedia}(2025{\natexlab{a}})}]{wiki_entanglement}
\bibinfo{author}{\bibnamefont{Wikipedia}}, \emph{\bibinfo{title}{Quantum entanglement}} (\bibinfo{year}{2025}{\natexlab{a}}), \bibinfo{note}{retrieved March 14, 2025}, \urlprefix\url{https://en.wikipedia.org/wiki/Quantum_entanglement}.

\bibitem[{\citenamefont{Wikipedia}(2025{\natexlab{b}})}]{wiki_kafka}
\bibinfo{author}{\bibnamefont{Wikipedia}}, \emph{\bibinfo{title}{Franz kafka}} (\bibinfo{year}{2025}{\natexlab{b}}), \bibinfo{note}{retrieved March 14, 2025}, \urlprefix\url{https://en.wikipedia.org/wiki/Franz_Kafka#%22Kafkaesque%22}.

\bibitem[{\citenamefont{Wikipedia}(2025{\natexlab{c}})}]{wiki_crispr}
\bibinfo{author}{\bibnamefont{Wikipedia}}, \emph{\bibinfo{title}{Crispr}} (\bibinfo{year}{2025}{\natexlab{c}}), \bibinfo{note}{retrieved March 14, 2025}, \urlprefix\url{https://en.wikipedia.org/wiki/CRISPR}.

\bibitem[{\citenamefont{Wikipedia}(2025{\natexlab{d}})}]{wiki_darkmatter}
\bibinfo{author}{\bibnamefont{Wikipedia}}, \emph{\bibinfo{title}{Dark matter}} (\bibinfo{year}{2025}{\natexlab{d}}), \bibinfo{note}{retrieved March 14, 2025}, \urlprefix\url{https://en.wikipedia.org/wiki/Dark_matter}.

\bibitem[{\citenamefont{Wikipedia}(2025{\natexlab{e}})}]{wiki_polo}
\bibinfo{author}{\bibnamefont{Wikipedia}}, \emph{\bibinfo{title}{Marco polo – the journey}} (\bibinfo{year}{2025}{\natexlab{e}}), \bibinfo{note}{retrieved March 14, 2025}, \urlprefix\url{https://en.wikipedia.org/wiki/Marco_Polo_%E2%80%93_The_Journey}.

\end{thebibliography}
\bibliographystyle{apsrev}

\appendix

\section{Proof of Equation~\ref{eq:averaged}}\label{appendix:d}

One should realize that, as stated above, taking $\hat{t}$ as the axis, the vectors $n_\alpha$ are uniformly distributed in the $d$-dimensional space. Thus, the angle $\theta_\alpha$ between $n_\alpha$ and $\hat{t}$ follows the distribution $f_\alpha(\theta_\alpha)$, which is given by:
\begin{align}
    f_\alpha(\theta_\alpha) \propto \sin^{d-2}{(\theta_\alpha)}\,.
\end{align}
This distribution assumes independence between vector $\vec{t}$ and angular projections' normal direction $n_\alpha$, as the complexity of the Transformer operator $\mathcal{T}$ precludes any weight-independent relationships between two vectors.

From this, it is straightforward to derive the expectation value of $\cos^2{\theta_\alpha}$ as:
\begin{align}
    \langle\cos^2{\theta_\alpha}\rangle = \frac{\int_0^\pi \sin^{d-2}{(\theta_\alpha)} \cos^2{\theta_\alpha} \, d\theta_\alpha}{\int_0^\pi \sin^{d-2}{(\theta_\alpha)} \, d\theta_\alpha}\,.
\end{align}

Now, recall that $d \gg 1$. To simplify the integral, we perform a change of variables by defining:
\begin{align}
    \theta_\alpha = \frac{\pi}{2} \pm \sqrt{\frac{2x}{d-1}}\,.
\end{align}

Under this transformation, the expectation value becomes:
\begin{align}
    \langle\cos^2{\theta_\alpha}\rangle = \frac{2\Gamma\left(\frac{3}{2}\right)}{(d-1)\Gamma\left(\frac{1}{2}\right)} = \frac{1}{d-1}\,,
\end{align}
where we have used the properties of the Gamma function, $\Gamma\left(\frac{3}{2}\right) = \frac{\sqrt{\pi}}{2}$ and $\Gamma\left(\frac{1}{2}\right) = \sqrt{\pi}$.

Finally, summing over all $\alpha$, we obtain:
\begin{equation}
    \sum_{\alpha} \langle \cos^{2}{\theta_{k,\alpha}} \rangle = \frac{\sum_{\alpha} 1}{d-1} = -\frac{\Delta d}{d-1}\,,
\end{equation}
where $\Delta d$ represents the reduced dimensions.

\section{Appendix: Wikipedia-Based Text Samples for Diversity Evaluation}
\label{appendix:wiki_samples}

We provide five text samples extracted from and modified based on Wikipedia to evaluate the model's cross-domain diversity. Each sample was curated from distinct Wikipedia entries to ensure coverage of the following domains: \textit{physics}, \textit{literature}, \textit{biology}, \textit{cosmology}, and \textit{history}. The selection criteria prioritized:  
\begin{itemize}
    \item Domain Coverage: Maximizing representativeness across STEM and humanities
    \item Stylistic Variation: Balancing technical definitions (e.g., CRISPR mechanisms), analytical discourse (e.g., Kafkaesque interpretations), and historical narratives (e.g., Marco Polo's travels)
\end{itemize}

\subsection{Sample 1: Scientific Explanation}
\label{subsec:1}
Domain: Quantum Physics~\cite{wiki_entanglement}\\
Total Length: 385 words\\
Content: Quantum entanglement is the phenomenon of a group of particles being generated, 

\textit{(The following details are omitted for brevity.)}

The use of quantum entanglement in communication and computation is an active area of research and development. You

\subsection{Sample 2: Literary Analysis}
Domain: Literary Criticism~\cite{wiki_kafka}\\
Total Length: 270 words\\
Content: "Kafkaesque" redirects here. For the Breaking Bad episode, see Kafkaesque (Breaking Bad).
The term "Kafkaesque" is used to describe concepts and situations reminiscent of Kafka's work, particularly Der Prozess (The Trial) and Die Verwandlung (The Metamorphosis).

\textit{(The following details are omitted for brevity.)}

paraphrased in "What it Means to be Kafkaesque" by Joe Fassler in The Atlantic, "Kafka's quintessential qualities are affecting use of language, a setting that straddles fantasy and reality, and a sense of striving even in the face of bleakness—hopelessly and full of hope."

\subsection{Sample 3: Biotechnology Overview}
Domain: Molecular Biology~\cite{wiki_crispr}\\
Total Length: 629 words\\
Content: CRISPR (an acronym for clustered regularly interspaced short palindromic repeats) is a family of DNA sequences found in the genomes of prokaryotic organisms such as bacteria and archaea.

\textit{(The following details are omitted for brevity.)}

 thermophilus strains for yogurt production. Danisco was later bought by DuPont, which owns about 50 percent of the global dairy culture market, and the technology spread widely.

\subsection{Sample 4: Cosmology Review}
Domain: Astrophysics~\cite{wiki_darkmatter}\\
Total Length: 416 words\\
Content: In astronomy, dark matter is an invisible and hypothetical form of matter that does not interact with light or other electromagnetic radiation. Dark matter is implied by gravitational effects which cannot be explained by general relativity unless more matter is present than can be observed.

\textit{(The following details are omitted for brevity.)}

These include modified Newtonian dynamics, tensor–vector–scalar gravity, or entropic gravity. So far none of the proposed modified gravity theories can describe every piece of observational evidence at the same time, suggesting that even if gravity has to be modified, some form of dark matter will still be required.
    
\subsection{Sample 5: Historical Narrative}
Domain: Medieval History \& Ethnomusicology~\cite{wiki_polo}\\
Total Length: 377 words\\
Content:The 13th century was the time of great political and economic expansion in the Republic of Venice. Marco Polo was a famous Venetian merchant traveler whose travels are recorded in Il Milione, a book which did much to introduce Europeans to Central Asia and China. 

\textit{(The following details are omitted for brevity.)}

As an important civil servant, Marco Polo was able to partake in the ceremonies of the court and thus listen to the ceremonial music played by huge orchestras with 100 or more musicians.

\end{document}